\documentclass[10pt,journal,twoside,compsoc]{IEEEtran}
\pdfoutput=1
\usepackage{amssymb}
\usepackage{gensymb}
\usepackage{stmaryrd}
\usepackage{cite}
\usepackage{color}
\usepackage{multirow}
\usepackage{graphicx,times}
\usepackage{epstopdf}
\usepackage{indentfirst}
\usepackage{CJK}
\usepackage{amsmath}
\usepackage{amsfonts}
\usepackage{txfonts}
\usepackage{mathrsfs}
\usepackage{theorem}
\usepackage{float}
\usepackage{subfigure}

\usepackage{verbatim}
\usepackage{hhline}     		% extended styles for tables
\usepackage{algorithmic}
\usepackage{algorithm}
\usepackage{array}
\usepackage{url}
\usepackage{bm}
\usepackage{xcolor}
\usepackage{microtype}
\usepackage{cleveref}
\usepackage{balance}

\crefname{section}{\S}{\S\S}
\Crefname{section}{\S}{\S\S}

\newcommand{\ie}{{\em i.e.}}
\newcommand{\eg}{{\em e.g.}}
\newcommand{\et}{{\em et al.}}

\newcommand\inv[1]{#1\raisebox{1.15ex}{$\scriptscriptstyle-\!1$}}

\def\spth{\textsuperscript{th}}

\begin{document}
\title{LoRa Backscatter Assisted State Estimator for Micro Aerial Vehicles with Online Initialization}

\author{\IEEEauthorblockN{Shengkai Zhang,~\IEEEmembership{Student Member,~IEEE,}
        Wei Wang,~\IEEEmembership{Senior Member,~IEEE,}
        Ning Zhang,
        and Tao Jiang,~\IEEEmembership{Fellow,~IEEE}}
\thanks{Part of this work has been presented at IEEE INFOCOM 2020~\cite{zhang2020rf}.}
\thanks{This work was supported in part by the National Key R\&D Program of China under Grant 2019YFB180003400, 2020YFB1806606, National Science Foundation of China with Grant 62071194, 91738202, Young Elite Scientists Sponsorship Program by CAST under Grant 2018QNRC001. \textit{ (Corresponding author: Wei Wang.) }}
\thanks{S. Zhang, W. Wang, N. Zhang, and T. Jiang are with the School of Electronic Information and Communications, Huazhong University of Science and Technology, Wuhan, Hubei, China.\protect\\
E-mail: \{szhangk, weiwangw, ning\_zhang, taojiang\}@hust.edu.cn.}
}

\markboth{IEEE TRANSACTIONS ON MOBILE COMPUTING,~Vol.~00, No.~00, 00~0000}
{Zhang {\em et al.}: LoRa Backscatter Assisted State Estimator for Micro Aerial Vehicles with Online Initialization}

\sloppy
\IEEEcompsoctitleabstractindextext{
\begin{abstract}
The advances in agile micro aerial vehicles (MAVs) have shown great potential in replacing humans for labor-intensive or dangerous indoor investigation, such as warehouse management and fire rescue. However, the design of a state estimation system that enables autonomous flight poses fundamental challenges in such dim or smoky environments. Current dominated computer-vision based solutions only work in well-lighted texture-rich environments. This paper addresses the challenge by proposing Marvel, an RF backscatter-based state estimation system with online initialization and calibration. Marvel is nonintrusive to commercial MAVs by attaching backscatter tags to their landing gears without internal hardware modifications, and works in a plug-and-play fashion with an automatic initialization module. Marvel is enabled by three new designs, a backscatter-based pose sensing module, an online initialization and calibration module, and a backscatter-inertial super-accuracy state estimation algorithm. We demonstrate our design by programming a commercial MAV to autonomously fly in different trajectories. The results show that Marvel supports navigation within a range of $50$ m or through three concrete walls, with an accuracy of $34$ cm for localization and $4.99\degree$ for orientation estimation. We further demonstrate our online initialization and calibration by comparing to the perfect initial parameter measurements from burdensome manual operations.
\end{abstract}

}

\maketitle

\IEEEdisplaynontitleabstractindextext

\IEEEpeerreviewmaketitle

\section{Introduction}
\label{sec:intro}
Over the last decade, the rapid proliferation of micro aerial vehicles (MAV) technologies has shown great potential in replacing human for labor-intensive or even dangerous indoor investigation and search, such as warehouse inventory management and fire rescue~\cite{lin2018autonomous, dhekne2019trackio, guo2019localization}. Specifically, using MAVs to manage inventory for warehouses cuts inventory checks from one month down to a single day~\cite{walmart_drone}, and using MAVs for search and rescue in firefighting operations saves the lives of firefighters by the fact that $53\%$ of deaths of the firefighters in the United States occurred in burning buildings in 2017~\cite{firefighter}. These applications require MAVs navigating autonomously in dim warehouses~\cite{fichtinger2015assessing} or smoky buildings while reporting to a server or terminal at a distance or through walls. 

{\em State estimation} is fundamental to the autonomous navigation of MAVs. The state, including position, velocity, and orientation, is the key to the flight control system of an aerial vehicle that adjusts the rotating speed of rotors to achieve desired actions for responding remote control or autonomous operations. The mainstream uses GPS, compass and vision sensors to estimate a MAV's state. However, GPS-compass based approaches~\cite{farrell2008aided, chao2010autopilots} only work in outdoor free space since GPS signals can be blocked by occlusions and compass measurements are easily distorted by surrounding environments. In indoors, computer vision (CV) based approaches have attracted much attentions due to their lightweight, high accuracy, and low cost, while limited to good lighting or texture-rich environments~\cite{zhu2017event, dong2019pair, qin2017vins, mur2015orb, dong2018vinav}, thereby failing to work in dim warehouses or smoky fire buildings. 

Recent years have witnessed much progress in using RF signals to track a target's pose (position and orientation), holding the potential to state estimation that is highly resilient to visual limitations. Despite novel systems that have led to high accuracy~\cite{liu2017cooperative, luo20193d, jiang20193d, wu2019sigcomm_rim, liu2017mercury, zhang2019self}, hardly any of these ideas have made it into the scenario of indoor MAVs that needs the following requirements:

\begin{itemize}
	\item {\bf Long range/through wall.} To scan items across a warehouse or navigate in a fire building, the system should support the navigation at least across rooms or over an area of tens of meters.
	\item {\bf Lightweight.} As a MAV is typically compact with limited battery capacity, it requires a lightweight, small-sized, and low-power sensing modality to enable state estimation. 
	\item {\bf Plug-and-play.} To make it practical to emergency rescue and efficient indoor investigation, the system should be instantly operational in an unknown environment. 
	\item {\bf Online initialization.} As a MAV state estimator is nonlinear and thus it is inoperable without a good initial state estimate, a convenient online initialization is required.
\end{itemize}

\begin{figure}[t]
  \centering
  \includegraphics[width=3.4in]{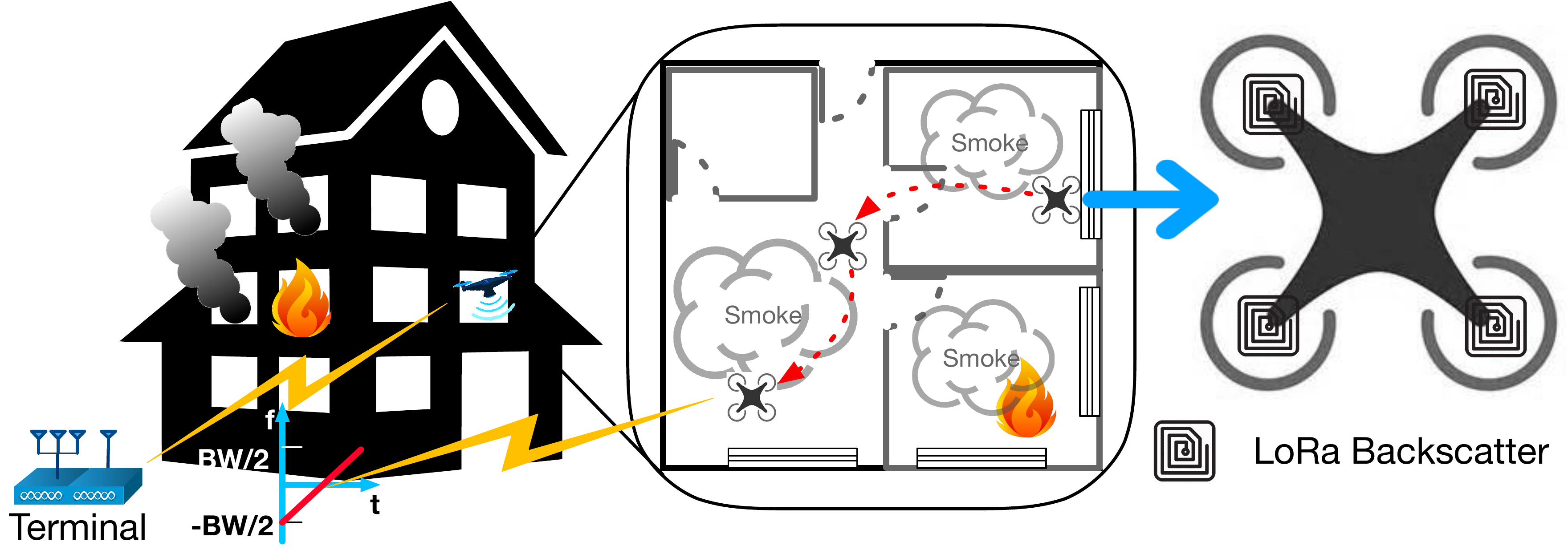}
  \caption{Usage example.}
  \label{fig:toy}
\end{figure}
\noindent Specifically, RFID-based solutions~\cite{luo20193d, jiang20193d} are lightweight while their operational range is limited. UWB and WiFi based solutions~\cite{liu2017cooperative, wu2019sigcomm_rim, zhang2019self} have better operational range while requiring pre-deploying multiple anchors to enable the localizability, failing to meet the plug-and-play requirement. And all these solutions lack online initialization methods when applying them on MAVs.

In this paper, we present Marvel, a state estimation system for MAVs that satisfies all these requirements. As shown in Fig.~\ref{fig:toy}, Marvel is able to support the navigation in burning buildings with smoke and fog in firefighting operations. There are four LoRa backscatter tags attached on the vehicle's landing gear. The MAV's terminal on the ground sends CSS signals to excite the tags and enable the state estimation, allowing the vehicle to fly across rooms or in an area of several tens of meters. The system is lightweight in that it merely attaches a few backscatter tags to the landing gear of a commercial MAV without any internal hardware modification. Marvel works in a plug-and-play fashion in that we augment the user's terminal of the MAV system with a LoRa signal processing module to enable state estimation, without adding any hardware. Moreover, Marvel does not need to the prior knowledge of the terminal's location. The terminal's position and the initial state can be initialized online by hand holding and moving the MAV around for few seconds before takeoff. The design of Marvel is structured around three components:

{\bf (a) Backscatter-based pose sensing:} Marvel's first component enables a backscatter-based sensing modality that allows it to estimate the response of the attached tags over backscattered signals that are drowned by noise. This sensing modality leverages chirp spread spectrum (CSS) signals to enable the pose tracking in long range or through occlusions where the signal amplitudes and phases are not available. It introduces a set of algorithms that first estimate channel phases of the tags' backscatter signals under mobility and then use the phases to estimate pose features, including the range, angle and rotation of the MAV to its terminal. This component enables Marvel to operate across rooms or in an area of several tens of meters. 
 
{\bf (b) Online initialization and extrinsic calibration:} Based on the estimates of the first component, we aim to accurately compute the state of a MAV. The problem will be a nonlinear optimization due to the vehicle's rotation. Thus, we need a good initialization point to bootstrap a nonlinear solver, \eg, the Gauss-Newton algorithm. Furthermore, since sensors provide measurements in their own reference frames, we require to calibrate the relative pose between the backscatter sensing and the IMU, \ie, {\em extrinsic parameters}, to align their measurements. We propose an approach for joint initialization and extrinsic calibration without any knowledge about the sensor layout. This online approach makes our design more practical to obtain a good initialization point and support various commercial products.

{\bf (c) Backscatter-inertial super-accuracy state estimation:} Marvel's third innovation is a backscatter-inertial super-accuracy algorithm, which is to solve a highly nonlinear problem that needs the initialization and extrinsic parameters from the previous component to bootstrap the solution. Our method combines the backscatter-based estimates with the onboard IMU measurements to enable accurate state estimation. Despite that IMU suffers in error accumulation, our backscatter sensing is drift-free, \ie, no temporal error accumulation, being able to correct the IMU drift by multi-sensor fusion. It employs a graph-based optimization framework to compute a Gaussian approximation of the posterior over the MAV trajectory. This involves computing the mean of this Gaussian as the configuration of the state that maximizes the likelihood of the sensor observations.

{\bf Results}. We build a prototype of Marvel using a DJI M100 MAV attached with four LoRa backscatters customized by off-the-shelf components. We demonstrate Marvel's potential by programming the MAV to autonomously fly in different trajectories in a long-range open space and an across-room indoor test site. The results show that Marvel can support the navigation within a range of $50$ m or with three concrete walls blocked the vehicle to its terminal and achieves an average accuracy of $34$ cm for localization and $4.99\degree$ for orientation estimation. In addition, we further demonstrate the effectiveness of our online initialization and calibration by exhibiting a similar performance to the perfect initial parameter measurements from burdensome manual operations.

{\bf Contributions}. We introduce a novel backscatter-based pose sensing technique that first extracts the channel phases of CSS signals under mobility and then estimates pose features of a MAV by the phases. This enables pose tracking in long range or through occlusions. Second, we propose an online initialization and extrinsic calibration approach that makes Marvel more practical. Furthermore, we design a backscatter-inertial super-accuracy algorithm that fuses backscatter-based estimates and IMU measurements to enable accurate state estimation. Finally, we implement a prototype and conduct real-world experiments demonstrating the system's ability to navigate a MAV across rooms or in an area of several tens of meters.

The rest of this paper is organized as follows. We first introduce the background of autonomous  flight and give an overview of our system in \cref{sec:overview}. Then we elaborate on the three designs: the backscatter pose sensing in \cref{sec:lr}, the online initialization and extrinsic calibration in \cref{sec:init}, and the super-accuracy state estimation in \cref{sec:pose}. The system implementation and performance evaluation are detailed in \cref{sec:evaluation}. We review related works in \cref{sec:related} and \cref{sec:conclusion} concludes our work.

\section{Background and System Overview}
\label{sec:overview}
\subsection{Background}
\begin{figure}[t]
  \centering
  \includegraphics[width=3.5in]{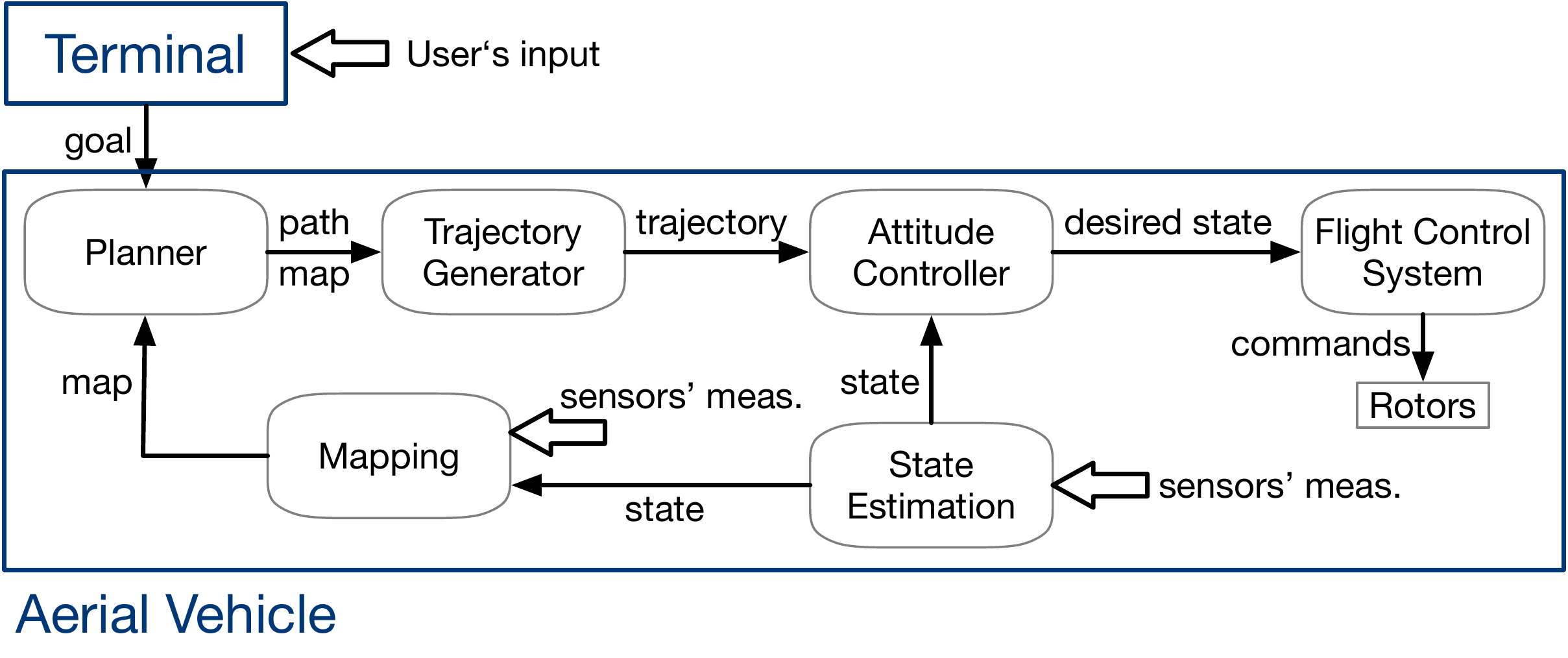}
  \caption{A framework of autonomous aerial vehicles.}
  \label{fig:background}
  \vspace{-4mm}
\end{figure}

Fig.~\ref{fig:background} shows a typical framework of autonomous aerial vehicles. A terminal on the ground is the user interface that sends desired goals to an aerial vehicle or multiple aerial vehicles. The level of the vehicle autonomy determines the types of user goals that can be supported. Semi-autonomous flight allows users to give desired actions, \eg, forward, backward, and turning left/right. Most of commercial photographic drones work in this way. A user is trained to control the vehicle's action by the remote terminal. On the other hand, fully autonomous flight can accept high-level goals, \eg, cruise on a certain track, requiring the integration of multiple technologies. 

Four components are required to enable fully autonomous flight~\cite{mohta2018fast}. The first component is state estimation. This refers to the ability of a vehicle to estimate its position, orientation, and velocity (the rate of change of position and orientation). Second, the vehicle must be able to compute its control commands. Based on where it needs to go and what the estimate of its current state is, the vehicle must be able to compute the commands that need to be sent to the motors or the rotors, and have them rotate at the appropriate speeds to achieve the desired action. Third, the vehicle needs some basic capability to map its environment. If it does not know what the surrounding looks like, then it is incapable of reasoning and planning safe trajectories in this environment. Finally, the vehicle should be able to compute safe paths, given a set of obstacles and a destination. 

The terminal sends a desired goal to the planner of the vehicle to start a task. The planner generates a path using the map from the mapping module. It then sends the path to the trajectory generator. The trajectory generator converts the path into a trajectory and sends it to the attitude controller. The attitude controller derives the desired state based on the trajectory and the current estimated state. It then sends the desired state to the flight control system. The flight control system takes the desired state to compute control commands to adjust the rotation speed of rotors. The input to the mapping module and the state estimation module denotes sensor measurements. This paper focuses on the fundamental and foremost important module -- state estimation.

\subsection{System Overview}
\begin{figure*}[htp]
  \centering
  \includegraphics[width=7in]{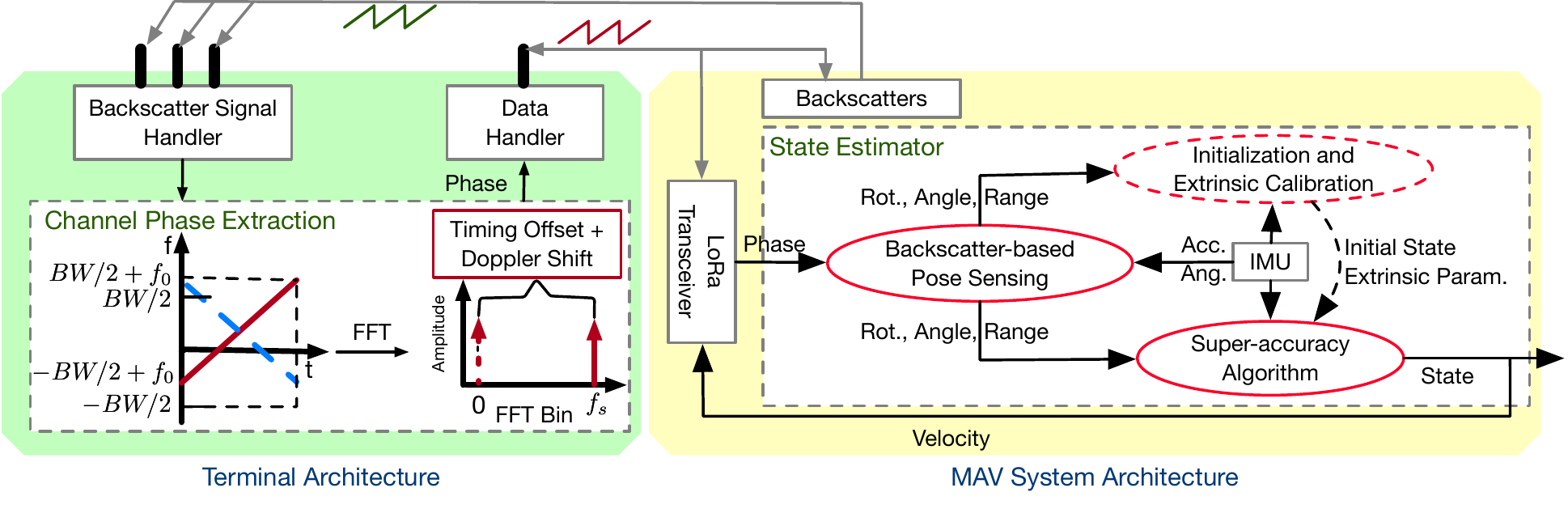}
  \caption{System architecture. The terminal receives backscattered signals and extracts their phases. The MAV system runs three components. Among them, the system only runs the initialization and extrinsic calibration at the initialization stage for bootstrapping the super-accuracy algorithm.}
  \label{fig:overview}
\end{figure*}

The system overview is shown in Fig.~\ref{fig:overview}. It has two components, the user's terminal and the MAV system. 

The terminal excites the backscatter tags on the MAV's landing gear and extracts channel phases. It has four antennas. We take one antenna as the data handler that alternately sends dummy chirps and data, such as channel phases. Dummy chirps are to excite the backscatter tags. Data are received by the LoRa transceiver on the MAV for pose sensing. We take the other three antennas as the backscatter signal handler that receives the signals backscattered by the tags and extracts the channel phases. The core module in the terminal is the {\em channel phase extraction} (see details in \cref{subsec:phase}), which provides channel phases for the backscatter-based pose sensing running on the MAV.

The MAV system runs the state estimator that takes channel phases from backscattered signals and measurements from the onboard IMU to estimate the state. The estimator consists of three modules: 
\begin{itemize}
	\item {\bf Backscatter-based pose sensing} uses the phases from the LoRa transceiver to compute the range, angle and rotation of the MAV to the terminal, enabling pose tracking in long range or through occlusions (see details in \cref{subsec:pose}).
	\item {\bf Online initialization and extrinsic calibration} takes the above backscatter-based pose estimates and the IMU measurements, which include 3D accelerations and angular velocities, to estimate the initial state and the extrinsic parameters, \ie, the relative pose between the backscatter sensing and the IMU (see details in \cref{sec:init}). The obtained initialization point and extrinsic parameters properly bootstrap the state estimation algorithm in the next component.
	\item {\bf Backscatter-inertial super-accuracy state estimation algorithm} fuses the measurements from the backscatter-based sensing and the IMU through a graph-based optimization framework. It models each module's estimates as a Gaussian mixture and computes a Gaussian approximation of the posterior over the MAV trajectory. The algorithm updates the state in a sliding window fashion for real-time processing (see details in \cref{sec:pose}).
\end{itemize}

Finally, the state estimator sends the state to the MAV's flight control system, \ie, DJI N1 flight control system~\cite{djim100} in our implementation. With the current estimated state and a desired goal, the flight control system computes the commands that adjust the power to rotors to achieve desired actions.

\section{Backscatter-based Pose Sensing}
\label{sec:lr}
\subsection{Primer on CSS Processing}
\label{subsec:primer}
In our system, the terminal transmits a linear upchirp signal with a bandwidth $BW$ to the MAV with backscatter tags. A tag backscatters the signal with a frequency shift $f_0$ for preventing the interference from the excitation signal. Multiple tags have different frequency shifts. Marvel uses the chirp signal that is compatible with LoRa protocol. We adopt the configuration options provided by Semtech LoRa chipset~\cite{sx1276}. It sets a fixed number of options for each parameter, \eg, $SF \in \{6, 7, 8, 9, 10, 11, 12\}$, with the recommendations for using these parameters. Its duration $T$ depends on spreading factor ($SF$) and bandwidth~\cite{liando2019known, note2015loratm}, \ie, $T = 2^{SF}/BW$. The most prominent recommendation is to use $SF$ settings of $SF = 7$ to $12$ and $BW$ $125$, $250$, and $500$ KHz. The recommendations ensure the acceptable transmission distance and data rate tradeoff. 

In our context, to achieve state estimation in a long-range/through-wall setting, the CSS signal needs good decoding capability. This capability is proportional to the product of signal duration $T$ and bandwidth $BW$. As $T\times BW = 2^{SF}$, we choose $SF = 12$. Meanwhile, to improve the range resolution of the backscatter-based pose sensing (\cref{subsec:pose}), we need the signal bandwidth as large as possible. Thus, we set $BW = 500$ KHz. At this configuration, the chirp duration is $8.192$ ms. Such a short chirp duration, which is within the channel coherent time, is required by the channel phase extraction (\cref{subsec:phase}).

To decode the chirp, the receiver first multiples the received signal with a synthesized downchirp whose frequency linearly varies from $BW/2 + f_0$ to $-BW/2 + f_0$. Then, it takes a fast Fourier transform (FFT) on this multiplication (Fig.~\ref{fig:overview}). This operation sums the energy across all the frequencies of the chirp, producing a {\em peak} at an FTT bin. 

\subsection{Below-Noise Channel Phase Extraction}
\label{subsec:phase}
Since MAVs are expect to carry out emergency tasks like fire rescue, the system desires the localizability with a single anchor (its terminal) and without prior knowledge of the work space, being instantly deployable and operable wherever required. The position of a target referring to a single anchor can be represented by the angle $\phi$ and the range $r$ of the target to the anchor as polar coordinates. And both the parameters can be inferred by the channel phase of the signal. 

The channel phase extraction for chirp signals has two steps as shown in Fig.~\ref{fig:phase_workflow}: we first combat the Doppler effect to estimate the beginning of the chirp and then we extract the channel phase leveraging the linearity of the chirp frequencies. 

To estimate the beginning of the chirp, we leverage a key property of the chirp signal: a time delay in the chirp signal translates to frequency shift. Ideally, decoding the original upchirp with a downchirp produces a peak in the first FFT bin (see Fig.~\ref{fig:overview}). When a tag is separated from the terminal, the backscatter signal handler receives the signal with a timing offset of the signal's round trip. The peak appears in the shifted bin $f_s$. If we move the beginning of the received chirp $f_s$ samples closer to its real beginning and repeat the decoding operation, there will be a new peak at the first FFT bin again and the symbol at this instant is the beginning of the transmission. However, under the MAV's mobility, the signal additionally experiences the Doppler frequency shift. The shifted bin $f_s$ is a mixed result of the timing offset and the Doppler effect. The above operation can no longer recover the beginning of the chirp.

Our solution leverages the kinetics and the structure of a MAV. We attach four backscatters on the landing gear of a MAV. As shown in Fig.~\ref{fig:tags}, the Doppler frequency shift of a tag, \eg, tag $T_1$, is a combinatorial result of translation and rotation. The shift $\Delta f(t)$ can be expressed as, 
\begin{equation}
	\Delta f(t) = \frac{f_c}{c} \mathbf{u}_p(t)\cdot\left[\mathbf{v}_t(t) + \mathbf{v}_r(t)\right] = \Delta f_t(t) + \Delta f_r(t),
	\label{eqn:doppler}
\end{equation}
where $\mathbf{u}_p(t)$ is the unit vector that represents the direction from the MAV to the terminal, $f_c$ the carrier frequency, $c$ the speed of RF signals in the medium. $\mathbf{v}_t(t)$ and $\mathbf{v}_r(t)$ are the translational velocity and the rotational velocity. $\Delta f_t(t)$ and $\Delta f_r(t)$ corresponds to the translational shift and the rotational shift. To estimate the beginning of the chirp, we need to isolate the frequency shift translated from the timing offset by eliminating the effect of Doppler shift. 

\begin{figure}
	\centering
	\begin{minipage}[b]{0.22\textwidth}\centering
		\center
		\includegraphics[width=1\textwidth]{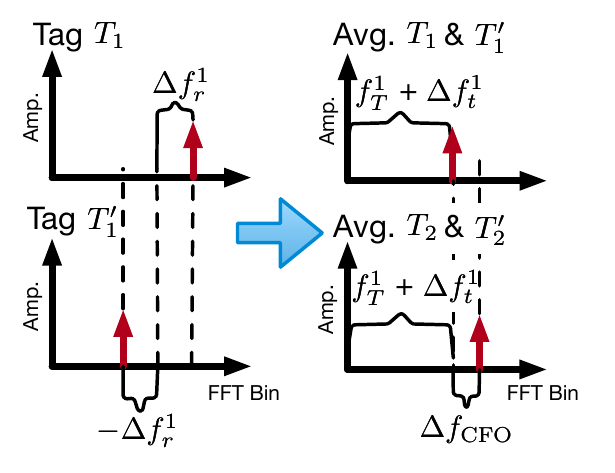}\vspace{-0.3cm}
		\caption{Rotational shift elimination.} \label{fig:peaks}
	\end{minipage}
	\hspace{0.1cm}
	\begin{minipage}[b]{0.24\textwidth}\centering
		\center
		\includegraphics[width=1\textwidth]{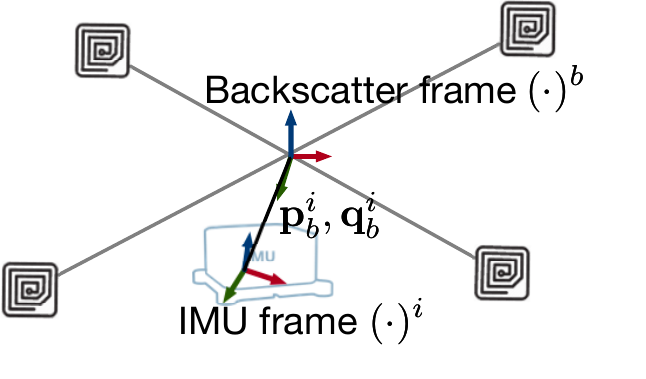}\vspace{-0.3cm}
		\caption{The reference frames of two sensing components in our system.} \label{fig:extrinsic}
	\end{minipage}
\end{figure}

{\bf Eliminating the effect of Doppler shift}. 
We first eliminate the effect of the rotational shift by the key observation that any pair of opposing tags on the landing gear, \eg, tags $T_1$ and $T_1^\prime$ in Fig.~\ref{fig:tags}, always have rotational velocities with {\em the same magnitude but opposite directions} and all tags share {\em the same translational velocity}. Thus, averaging the shifted peak of two opposing tags eliminates the rotational shift as shown in Fig.~\ref{fig:peaks}. Specifically, decoding the backscattered signals from a pair of opposing tags, we obtain the FFT bin indices, $\hat{B}_i$ and $\hat{B}_i^\prime$,
\begin{equation}
	\hat{B}_i = f_T^i + \Delta f_t^i + \Delta f_r^i, \; \hat{B}_i^\prime = f_T^{i^\prime} + \Delta f_t^i - \Delta f_r^i,
	\label{eqn:shift}
\end{equation}
where $f_T^i$ and $f_T^{i^\prime}$ are the frequency shift translated by the timing offset, $\Delta f_t^i$ and $\Delta f_r^i$ the translation shift and the rotational shift of tag $i$. Note that $f_T^i \approx f_T^{i^\prime}$ since their maximal difference is the translated shift from the traveling time of the distance between a pair of opposing tags, \ie, the diameter $D$ of the MAV, which is negligible as $D$ ($66$ cm for the DJI M100) is too small for the speed of RF signal propagation. Thus, averaging them, \ie, $1/2(\hat{B}_i + \hat{B}_i^\prime) = f_T^i + \Delta f_t^i$, eliminates the rotational shift.

Since the two pairs of tags on the MAV are structurally symmetric, when we perform the above operation to each pair, the results are expected to be identical. However, they exhibit a slight difference as shown in Fig.~\ref{fig:peaks}. This is because the micro-controllers of the tags are not synchronized with the terminal, it introduces an additional carrier frequency offset (CFO) for each tag, which is a constant. In our approach, $\Delta f_{\text{CFO}}$ is the difference of CFOs upon averaging the two pairs of tags, which is still a constant. We can simply apply this to the rest of the transmission to estimate the right chirp phase.

Now we eliminate the translational shift $\Delta f_t^i$ to isolate the frequency shift $f_T^i$ translated from the timing offset. Then, we can obtain the signal at the real beginning of the transmission by moving the beginning of the received chirp $f_T^i$ samples. $\Delta f_t^i$ can be tracked using the accelerations measured by the onboard IMU. Initially, the MAV is about to take off. At this initial stage, there is no motion, $1/2(\hat{B}_i + \hat{B}_i^\prime)$ is already the frequency shift $f_T^i$. Thus, the channel phase can be obtained according to the workflow (Fig.~\ref{fig:phase_workflow}). Then, we specify $\mathbf{u}_p(t)$ in Eqn.~\eqref{eqn:doppler} by our angle estimation algorithm in \cref{subsec:pose}. When the MAV takes off, the accelerations measured by IMU can track the translational velocity $\mathbf{v}_t(t)$. Thus, $\Delta f_t^i = f_c/c\cdot\mathbf{u}_p(t)\cdot\mathbf{v}_t(t)$ and $f_T^i = 1/2(\hat{B}_i + \hat{B}_i^\prime) - \Delta f_t^i$. Note that integrating the accelerations to obtain the velocity will suffer from the temporal drift. The super-accuracy algorithm in \cref{sec:pose} corrects the drift and feeds back to the flight control system.

{\bf Extracting channel phase}.
At this stage, we have corrected the signal to the symbol at the beginning of the transmission. Now we compute the channel phases of all frequencies in the chirp by the method proposed in~\cite{nandakumar20183d}. We have
\begin{equation}
\hat{\theta}_\Sigma = \theta_1 + \theta_2 + \cdots + \theta_N = \theta_1 + \theta_1 \frac{f_2}{f_1} + \cdots + \theta_1 \frac{f_N}{f_1},
	\label{eqn:phase_sum}
\end{equation}
where $f_1, \cdots, f_N$ are explicitly defined when generating the chirp signal. Solving the above equation obtains the channel phases of all frequencies in the chirp.

Notice that this method requires a short chirp duration to be within the channel coherent time. As we mentioned in \cref{subsec:primer}, we choose the parameters of CSS signals that conform to LoRa standard as $SF = 12$, $BW = 500$ KHz. According to signal duration $T = 2^{SF}/BW$, the chirp duration is $8$ ms, which is within the channel coherent time. Moreover, $SF = 12$ ensures the best decoding capability of CSS signals and $BW = 500$ KHz will benefit the range estimation of the pose sensing in the next subsection.

\subsection{Below-Noise Pose Sensing}
\label{subsec:pose}
{\bf Range estimation}. 
Assume that the terminal is separated from a tag on the MAV by a distance of $r$. A linear chirp signal with $N$ frequencies transmitted by the terminal propagates a total distance of $2r$ for the round trip to and from the tag. The wireless channel of such a signal is, $\mathbf{H} = \left[\gamma_1 e^{-j2\pi f_1\frac{2r}{c}}, \gamma_2 e^{-j2\pi f_2\frac{2r}{c}}, \cdots, \gamma_N e^{-j2\pi f_N\frac{2r}{c}}\right]$, where $\gamma_i$ is the attenuation corresponding to frequency $f_i$ in the chirp, $i = \{1, \cdots, N\}$. In the absence of multipath, we can use the obtained channel phases of the backscatter signal to estimate the range $r$. However, due to multipath, the obtained phases is actually the sum of phases of the direct-path signal and the multipath-reflected signals.

To combat multipath while conforming to LoRa protocol, we dynamically send multiple chirps in the channels of $900$ MHz band and combine the phase information across all these channels to simulate a wideband transmission. At a high level, a wideband signal can be used to disambiguate the multipath. There are $13$ channels separated by $2.16$ MHz with respect to the adjacent channels. We have four tags on the MAV which are configured to different frequency shifts for preventing the interference from the excitation signal. So, the terminal can transmit excitation signals in $2$ channels and receive backscatter signals across $8$ channels. By combining them, the terminal sends the phases at all the channels to the MAV through LoRa. Then, the MAV computes the range estimate by using an inverse FFT on the phases to get the time-domain multipath profile. We use a fixed energy threshold over this profile to identify the closest (most direct) path from the MAV. 

{\bf Angle estimation}. 
The angle of incident signals $\phi$ is also encoded in the phases of the signals. The backscattered chirp signal received by a linear array with $M$ antennas from $K$ propagation paths has the measurement matrix $\mathbf{X}$, 
\begin{equation}
	\begin{aligned}
		& \mathbf{X}  = \left[\mathbf{x}_1 \dotsc \mathbf{x}_N \right] = \mathbf{S}\left[\mathbf{F}_1 \dotsc \mathbf{F}_N \right], \\
		& \mathbf{S}\mathbf{F}_i = \left[ \mathbf{s}(\phi_1) \dotsc \mathbf{s}(\phi_K) \right]\left[ \gamma_{i1} \dotsc \gamma_{iK} \right]^\top, i = \{1, \cdots, N\},	\\
		& \mathbf{s}(\phi_k) = \left[ 1 \; e^{-j\eta \sin(\phi_k)} \dotsc e^{-j(M-1)\eta \sin(\phi_k)}  \right]^\top,
	\end{aligned}
	\label{eqn:angle}
\end{equation}
where $k = \{1, \cdots, K\}$, $\mathbf{F}_i$ denotes the attenuation factors of $K$ paths at frequency $i$ in the chirp, $\gamma_{ij}$ the attenuation factor of path $j$ at frequency $i$. $\mathbf{S}$ is the steering matrix where $s(\phi_k)$ denotes the steering vector of path $k$, and the constant $\eta = 2\pi d\frac{f_c}{c}$ where $d$ is the antenna spacing. $\phi_k$ is the angle of interest. We can see that the angle only exists in the steering matrix, contributing the phases in the complex elements of matrix $\mathbf{X}$. 

\begin{figure}
	\centering
	\begin{minipage}[b]{0.23\textwidth}\centering
		\center
		\includegraphics[width=1\textwidth]{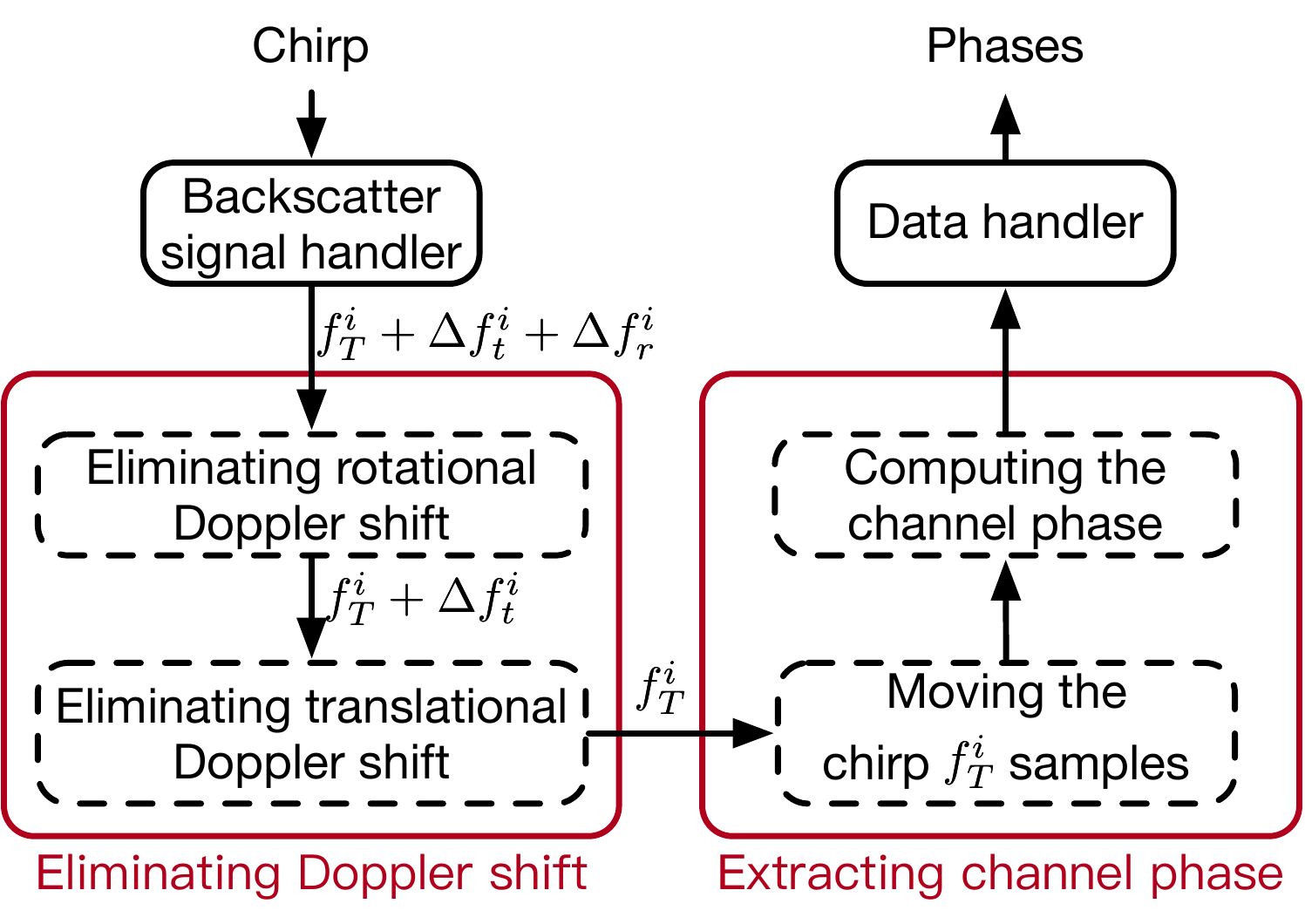}\vspace{-0.3cm}
		\caption{Phase extraction workflow.} \label{fig:phase_workflow}
	\end{minipage}
	\hspace{0.1cm}
	\begin{minipage}[b]{0.23\textwidth}\centering
		\center
		\includegraphics[width=1\textwidth]{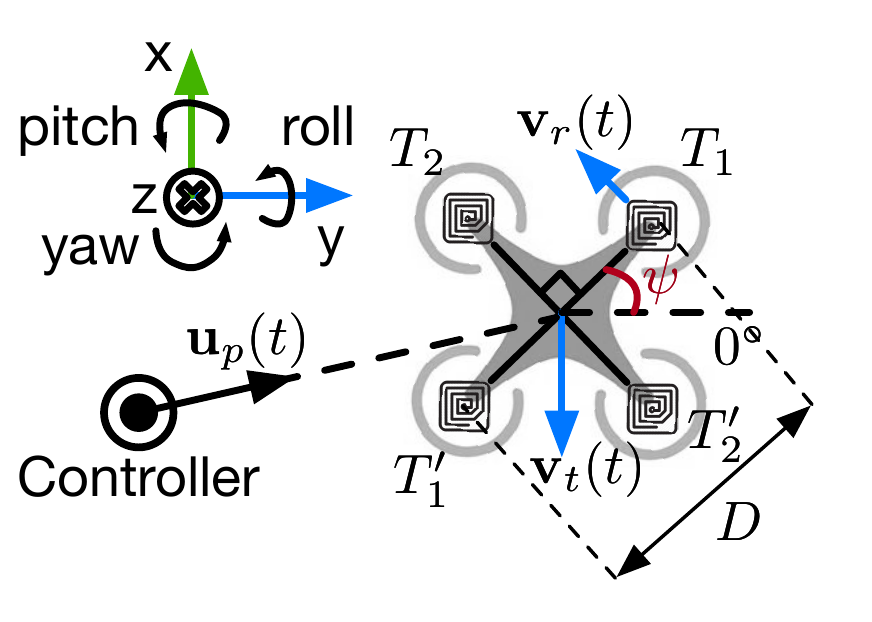}\vspace{-0.3cm}
		\caption{The motion of a MAV consists of translation and rotation.} \label{fig:tags}
	\end{minipage}
	\vspace{-4mm}
\end{figure}

Thus, even without the attenuation information, we can use the obtained phases to construct a {\em virtual measurement matrix} of which all complex elements have unit attenuation with the phases of frequencies in the chirp to allow the angle estimation. The virtual measurement matrix $\hat{\mathbf{X}}$ can be written as
\begin{equation}
	\hat{\mathbf{X}} = 
	\begin{bmatrix}
		e^{j\theta_{11}} 	& e^{j\theta_{12}} 	& \cdots 	& e^{j\theta_{1N}}	\\
		e^{j\theta_{21}} 	& e^{j\theta_{22}} 	& \cdots 	& e^{j\theta_{2N}}	\\
		\vdots				& \vdots 				& \ddots	& \vdots				\\
		e^{j\theta_{M1}} & e^{j\theta_{M2}}	& \cdots 	& e^{j\theta_{MN}}
	\end{bmatrix},
\end{equation}
where $\theta_{ij}$ denotes the phase of antenna $i$ at frequency $j$. Applying $\hat{\mathbf{X}}$ to the super-resolution angle estimation technique~\cite{kotaru2015spotfi}, we obtain the direct-path angle of a tag to the terminal. The four tags provide four angles for every chirp. We compute the harmonic mean of the four angles as the final result. 

{\bf Rotation estimation}. 
The real problem to determine a MAV's orientation is how to anchor the yaw, a.k.a., heading. The orientation can be represented by Euler angles: roll $\alpha$, pitch $\beta$, and yaw $\psi$ for a rotation around $x$, $y$, and $z$ axes (Fig.~\ref{fig:tags}). And it can be computed by integrating the 3D angular velocity readings from the onboard IMU. The results however suffer from temporal drifts due to the inherent noise of IMU. The drifts of roll and pitch would tilt the vehicle and move away. They can be corrected by the position, which has been obtained by the above range and angle estimates, as it helps the MAV realize unintended translations. However, the drift of heading causes no translation but rotation. We need drift-free rotation estimates to fix the heading. 
 
Our idea is that the rotational shift is solely determined by the rotation. We can use it to map the rotation. According to Eqn.~\eqref{eqn:shift}, subtracting the indices of the peaks from two opposing tags $\hat{B}_i$ and $\hat{B}_i^\prime$ gives the rotational frequency shift,
\begin{equation}
	\Delta \hat{B}_i = \hat{B}_i - \hat{B}_i^\prime  = f_T^i - f_T^{i^\prime} + 2\times \Delta f_r^i \approx 2\times \Delta f_r^i.
	\label{eqn:rotational_shift}
\end{equation}

Now we model the rotational shift. We denote the angle of the MAV to its terminal as $\phi$ and the MAV's rotation as $\psi$ (refer to Fig.~\ref{fig:tags}), then $\mathbf{u}_p = \left[\cos\phi \; \sin\phi \right]^\top, \; \mathbf{v}_r = \frac{D}{2}\omega\left[ \cos(\psi + \frac{\pi}{2}) \; \sin(\psi + \frac{\pi}{2}) \right]^\top$, where $\omega$ is the angular velocity during the rotation. The rotational shift can be expressed as
\begin{equation}
	\Delta f_r^i = \frac{f_c}{c} \mathbf{u}_p\cdot \mathbf{v}_r = \frac{f_cD}{2c} \omega \times \sin\left(\phi - \psi\right).
	\label{eqn:relative_shift}
\end{equation}
The terminal computes $\Delta f_r^i$ by Eqn.~\eqref{eqn:rotational_shift} and sends it to the MAV. $\phi$ can be obtained by the angle estimation algorithm. The gyroscope in IMU measures angular velocity $\omega$. The rest parameters are known constants. Thus, rotation $\psi$ can be solved by Eqn.~\eqref{eqn:relative_shift}. 

\section{Extrinsic Calibration and Initialization}
\label{sec:init}
So far, we have discussed how Marvel measures the pose feature, including the range, angle and rotation of a MAV to its terminal, based on backscattered CSS signals. We intend to estimate the state of the MAV by fusing these pose features with IMU measurements. As we will see in \cref{sec:pose}, due to the rotation estimation, the system model is highly non-linear, we first need a good initialization to bootstrap the solution. Besides, the extrinsic calibration is required when dealing with measurements from multiple sensors. The problem comes from the fact that each sensor provides measurements in its own reference frame as shown in Fig.~\ref{fig:extrinsic}. The pose features measured by the backscatter-based sensing are referred to the intersection of the four tags while the IMU is not exactly installed and aligned on that point. In this section, we elaborate on our online initialization and calibration approach to recover all initial states, including the positions of the vehicle and its terminal, velocity, attitude (represented by the gravity in the IMU frame), and extrinsic parameters.

We denote $(\cdot)^b$ as the backscatter frame, $(\cdot)^i$ the IMU frame. There exists a relative pose, $\mathbf{q}$ and $\mathbf{p}$, between the two coordinate frames. $\mathbf{q} = [q_x, q_y, q_z, q_c]^\top \in \mathbb{R}^4$ is the Hamilton quaternion~\cite{trawny2005indirect} representation of the rotation. $\mathbf{q}_b^i$ and $\mathbf{p}_b^i$ denotes the 3-D rotation and position of the backscatter frame with respect to the IMU frame. We further denote $(\cdot)^{i_k}$ as the IMU frame while obtaining the $k$\spth set of pose features from the backscatter sensing. Note that the representations of the world frame, a.k.a., the earth's inertial frame, is better for user's understanding. It is however usually difficult to align the initial frame to the world frame. Fortunately, as introduced by \cite{shen2015tightly}, if we change the system's reference frame to the frame of the first pose sensing $i_0$, the dependency on the global pose is removed. Therefore, we transform the output of our system to be at $i_0$ frame, \ie, refer to the initial pose. In addition, we use $(\hat{\cdot})$ to denote noisy sensor measurements.

\subsection{Extrinsic Rotation Estimation}
\label{subsec:extrinsic}

The online calibration and initialization procedure requires sufficient rotations. In practice, we hold the MAV and manually rotate it for this procedure. The extrinsic calibration and the initialization can be formulated as two linear systems. We first estimate the relative rotation $\mathbf{q}_b^i$ by aligning two rotation sequences from the IMU and the backscatter sensing. Then we use this rotation to further estimate the relative position $\mathbf{p}_b^i$ as well as other initial values in \cref{subsec:init}.

Typically, the IMU data rate ($100$ Hz) is much higher than the data rate of the backscatter sensing ($\approx 10$ Hz). Thus, there have been buffered multiple IMU measurements in the interval $[k, k+1]$. We first pre-integrate such IMU data between two sets of pose features from the backscatter sensing. The IMU pre-integration technique has been developed in~\cite{shen2015tightly, lupton2012visual}. We give its usage in our system.

The raw measurements of IMU includes acceleration $\hat{\mathbf{c}}_t$ and angular velocity $\hat{\bm{\omega}}_t$ at time $t$. Given two time instants that corresponds to two sets of backscatter-based pose features, we can pre-integrate the buffered IMU readings as~\cite{shen2015tightly}
\begin{equation}
	\begin{aligned}
		\hat{\bm{\alpha}}_{i_{k+1}}^{i_k} & = \iint_{t\in[k, k+1]}R(\mathbf{q}_t^{i_k})\hat{\mathbf{c}}_t\,\mathrm{d}t^2, \; \hat{\bm{\beta}}_{i_{k+1}}^{i_k} = \int_{t\in[k, k+1]}R(\mathbf{q}_t^{i_k})\hat{\mathbf{c}}_t\,\mathrm{d}t, \\
		\hat{\bm{\gamma}}_{i_{k+1}}^{i_k} & = \int_{t\in[k, k+1]}\bm{\gamma}_t^{i_k} \otimes \begin{bmatrix}
 							0 & \frac{1}{2}\hat{\bm{\omega}}_t
 						\end{bmatrix}^\top \mathrm{d}t,
	\end{aligned}
	\label{eqn:integration}
\end{equation}
where $\otimes$ denotes the quaternion multiplication operation. $R(\mathbf{q}_t^{i_k}) \in \text{SO}(3)$ is the conversion from the quaternion to the rotation matrix. We use the quaternion representation for modelling the odometry as a vector. $\bm{\gamma}_t^{i_k}$ is the incremental rotation from $i_k$ to current time $t$, which is available through short-term integration of gyroscope measurements. Then we can write the IMU propagation model for position, velocity, and rotation as~\cite{shen2015tightly}
\begin{equation}
  \begin{bmatrix}
  	 \mathbf{p}_{i_{k+1}}^{i_0}		\\
  	 \mathbf{v}_{i_{k+1}}^{i_{k+1}}		\\
  	 \mathbf{q}_{i_{k+1}}^{i_0}		\\
  \end{bmatrix} = 
  \begin{bmatrix}
  	\mathbf{p}_{i_k}^{i_0} + R(\mathbf{q}_{i_k}^{i_0})\mathbf{v}_{i_k}^{i_k}\Delta t_k - \frac{1}{2}\mathbf{g}^{i_0}\Delta t_k^2 + R(\mathbf{q}_{i_k}^{i_0})\hat{\bm{\alpha}}_{i_{k+1}}^{i_k}   \\
  	R(\mathbf{q}_{i_k}^{i_{k+1}})\mathbf{v}_{i_k}^{i_k} - R(\mathbf{q}_{i_0}^{i_{k+1}})\mathbf{g}^{i_0}\Delta t_k + R(\mathbf{q}_{i_k}^{i_{k+1}})\hat{\bm{\beta}}_{i_{k+1}}^{i_k}  \\
  	R(\mathbf{q}_{i_k}^{i_0})\hat{\bm{\gamma}}_{k+1}^k
  \end{bmatrix},
  \label{eqn:preimu}
\end{equation}
where $\Delta t_k$ denotes the time interval between two consecutive states. $\mathbf{g}^{i_0}$ is the initial gravity of the IMU frame. Note that IMU measurements combine the force for countering gravity $\mathbf{g}^{i_0}$ and the MAV dynamics. $\mathbf{g}^{i_0}$ is initially unknown since the initial attitude of a MAV is unknown. The gravity must be tracked to be aligned with the local frame through the attitude.

After obtaining the IMU rotation $\mathbf{q}_{i_{k+1}}^{i_k}$, we need to know the rotation $\mathbf{q}_{b_{k+1}}^{b_k}$ measured by the backscatter sensing. The two terms can be connected through the extrinsic rotation $\mathbf{q}_b^i$. 

The backscatter sensing estimates the yaw angle $\hat{\psi}_{b_{k+1}}^{b_k}$. Thanks to the IMU that it only drifts in four degrees of freedom, corresponding to 3D position and the yaw angle (rotation around the gravity direction)~\cite{qin2017vins}. In other words, the IMU has no drift in roll and pitch rotations. It is easy to obtain the incremental rotation from the backscatter sensing with the IMU reading. Specifically, we first compute roll $\hat{\vartheta}_{i_{k+1}}^{i_k}$ and pitch $\hat{\varphi}_{i_{k+1}}^{i_k}$ from $\hat{\bm{\gamma}}_{i_{k+1}}^{i_k}$ (Eqn.~\eqref{eqn:integration}). Note that the incremental roll and pitch in the two frames are identical in that the MAV is rigid, \ie, $\hat{\vartheta}_{b_{k+1}}^{b_k} = \hat{\vartheta}_{i_{k+1}}^{i_k}$ and $\hat{\varphi}_{b_{k+1}}^{b_k} = \hat{\varphi}_{i_{k+1}}^{i_k}$. Then we convert $(\hat{\vartheta}_{b_{k+1}}^{b_k}, \hat{\varphi}_{b_{k+1}}^{b_k}, \hat{\psi}_{b_{k+1}}^{b_k})$ to be $\mathbf{q}_{b_{k+1}}^{b_k}$.

With $\mathbf{q}_{i_{k+1}}^{i_k}$ and $\mathbf{q}_{b_{k+1}}^{b_k}$, $\mathbf{q}_{i_{k+1}}^{i_k} \otimes \mathbf{q}_b^i = \mathbf{q}_b^i \otimes \mathbf{q}_{b_{k+1}}^{b_k}$ holds for any $k$. Restructuring this equation gives
\begin{equation}
	\left[
		\mathcal{G}_1\left(\mathbf{q}_{i_{k+1}}^{i_k}\right) - \mathcal{G}_2\left(\mathbf{q}_{b_{k+1}}^{b_k}\right)
	\right] \cdot \mathbf{q}_b^i = \mathbf{G}_{k+1}^k \cdot \mathbf{q}_b^i = \mathbf{0},
\end{equation}
where
\begin{equation*}
	\begin{aligned}
		\mathcal{G}_1\left(\mathbf{q}\right) & = 
		\begin{bmatrix}
			q_c\mathbf{I}_3 + \lfloor\mathbf{q}_{xyz}\times\rfloor	&	\mathbf{q}_{xyz}	\\
			-\mathbf{q}_{xyz}		&	q_c
		\end{bmatrix}, \mathcal{G}_2\left(\mathbf{q}\right) = 
		\begin{bmatrix}
			q_c\mathbf{I}_3 - \lfloor\mathbf{q}_{xyz}\times\rfloor	&	\mathbf{q}_{xyz}	\\
			-\mathbf{q}_{xyz}		&	q_c
		\end{bmatrix}.
	\end{aligned}	
\end{equation*}
$\lfloor\mathbf{q}_{xyz}\times\rfloor$ is the skew-symmetric matrix from the first three elements $\mathbf{q}_{xyz}$ of the quaternion $\mathbf{q}$. $q_c$ is the fourth element.

With $N$ incremental rotations along the pose features from the backscatter sensing, we have the following over-constrained linear system
\begin{equation}
	\begin{bmatrix}
		\mathbf{G}_1^0	&	\mathbf{G}_2^1	&	\cdots	&	\mathbf{G}_N^{N-1}
	\end{bmatrix} \cdot \mathbf{q}_b^i = \mathbf{G}_N\cdot \mathbf{q}_b^i = \mathbf{0}.
\end{equation}
Solving the above system obtains the extrinsic rotation $\mathbf{q}_b^i$. Next, we take this to estimate the extrinsic translation $\mathbf{p}_b^i$ and the initial position, attitude, and velocity of the vehicle together.

\subsection{Initialization}
\label{subsec:init}

We adopt a sensor fusion method to obtain the initial state and employ a sliding window formation that incorporates a fixed number of IMU and backscatter sensing measurements to ensure constant computational complexity~\cite{kaess2012isam2}. We recover the initial state in the first IMU frame. The state vector within the window is defined as,
\begin{equation}
	\begin{aligned}
		\mathbf{\mathcal{S}} & = \left[ \mathbf{s}_0; \quad \mathbf{s}_1; \quad \cdots; \quad \mathbf{s}_n; \quad \mathbf{p}_b^i; \quad \bm{\rho} \right]		\\
		\mathbf{s}_k	& = \left[ \mathbf{p}_{i_k}^{i_0}; \quad \mathbf{v}_{i_k}^{i_k}; \quad \mathbf{g}^{i_k} \right], \; \mathbf{p}_{i_0}^{i_0} = 
		\begin{bmatrix}
			0	&	0	&	0
		\end{bmatrix}^\top,
	\end{aligned}
	\label{eqn:init}
\end{equation}
where $\mathbf{s}_k$ denotes $k$\spth state in the window, which contains position $\mathbf{p}_{i_k}^{i_0}$, velocity $\mathbf{v}_{i_k}^{i_k}$, and the gravity in the IMU frame $\mathbf{g}^{i_0}$, $n$ the number of states in the sliding window, $\bm{\rho}$ denotes the position of the terminal, $\mathbf{p}_b^i$ the relative translation of the IMU with respect to the backscatter sensing.

The initialization is to solve a maximum likelihood problem by minimizing the sum of the Mahalanobis norm of all measurements errors within the sliding window
\begin{equation}
	 \min_{\bm{\mathcal{S}}} \left\{\sum_{j\in\mathcal{L}}\left\| \hat{\mathbf{z}}_{b_j} - \mathbf{H}_{b_j}\bm{\mathcal{S}} \right\|_{\mathbf{P}_{b_j}}^2 + \sum_{k\in\mathcal{I}}\left\| \hat{\mathbf{z}}_{i_{k+1}}^{i_k} - \mathbf{H}_{i_{k+1}}^{i_k}\bm{\mathcal{S}} \right\|_{\mathbf{P}_{i_{k+1}}^{i_k}}^2\right\},
	 \label{eqn:linear}
\end{equation}
where $\mathcal{L}$ is the set of backscatter-based pose features and $\mathcal{I}$ denotes the set of IMU measurements. We choose the Mahalanobis norm to be the optimization objective because it takes into account the correlations of the data set. These correlations amongst internal states of different sensing modalities are key for any high-precision inertial-based autonomous system~\cite{leutenegger2015keyframe}. $\mathbf{H}_{b_j}$ and $\mathbf{H}_{i_{k+1}}^{i_k}$ are corresponding measurement matrices. Since the initialization procedure does not take long time, the gyroscope drift is not significant. We integrate the gyroscope measurements to compute rotation $\mathbf{q}_{i_{k+1}}^{i_0}$ and $\mathbf{q}_{i_{k+1}}^{i_k}$ and thus system \eqref{eqn:linear} can be solved in a linear fashion. 

We first define the measurement model $\left\{\hat{\mathbf{z}}_{b_j}, \mathbf{H}_{b_j}, \mathbf{P}_{b_j}\right\}$ for $j$\spth observation of backscatter sensing as
\begin{equation}
	\begin{aligned}
		\hat{\mathbf{z}}_{b_j} = \hat{\mathbf{0}} = \left\lfloor \left(\hat{d}_{b_j} \hat{\mathbf{a}}_{b_j}\right) \times \right\rfloor f_i^b\left(f_{i_0}^{i_j}\left(f_b^i\left(\mathbf{p}_{b_j} - \bm{\rho}\right)\right)\right) = \mathbf{H}_{b_j}\bm{\mathcal{S}} + \mathbf{n}_{b_j},
	\end{aligned}
\end{equation}
where $\hat{d}_{b_j}$ and $\hat{\mathbf{a}}_{b_j}$ are $j$\spth range and angle measurements from the backscatters. The function $f_X^Y(\mathbf{t})$ denotes the transformation of a vector $\mathbf{t}$ from frame $X$ to frame $Y$. We define $f_X^Y(\mathbf{t})$ and its inverse $f_Y^X(\mathbf{t})$ as
\begin{equation}
	\begin{aligned}
		f_X^Y(\mathbf{t}) & = R(\mathbf{q}_X^Y)\cdot\mathbf{t} + \mathbf{p}_X^Y, \quad f_Y^X(\mathbf{t}) = R(\mathbf{q}_Y^X)\cdot\left(\mathbf{t} - \mathbf{p}_X^Y\right).
	\end{aligned}
\end{equation}
Note that $f_{i_0}^{i_j}(\cdot)$ follows the same rule and all rotations are known. 

$\mathbf{n}_{b_j}$ is the additive Gaussian noise for the backscatter sensing. Its covariance matrix $\mathbf{P}_{b_j}$ can be estimated by statistically analyzing the pose features.

Then we can derive the IMU measurement model $\left\{\hat{\mathbf{z}}_{i_{k+1}}^{i_k}, \mathbf{H}_{i_{k+1}}^{i_k}, \mathbf{P}_{i_{k+1}}^{i_k}\right\}$ between consecutive frames $k$ and $k+1$ from Eqn.~\eqref{eqn:preimu} (with known $\mathbf{q}_{i_k}^{i_0}$ and $\mathbf{q}_{i_k}^{i_{k+1}}$) as~\cite{shen2016initialization} 
\begin{equation}
	\begin{aligned}
		\hat{\mathbf{z}}_{i_{k+1}}^{i_k} = 
		\begin{bmatrix}
			\hat{\bm{\alpha}}_{i_{k+1}}^{i_k}	\\
			\hat{\bm{\beta}}_{i_{k+1}}^{i_k}		\\
			\hat{\mathbf{0}}
		\end{bmatrix} & = 
		\mathbf{H}_{i_{k+1}}^{i_k} \bm{\mathcal{S}} + \mathbf{n}_{i_{k+1}}^{i_k},
	\end{aligned}
\end{equation}
where $n_{i_{k+1}}^{i_k}$ is the additional Gaussian noise for the IMU measurement model. The covariance $\mathbf{P}_{k+1}^k$ can be computed recursively by first-order discrete-time propagation within $\Delta t_k$, referring to~\cite{qin2017vins} for more details. Finally, we solve the above linear system to initialize the vehicle's state, extrinsic translation, and the terminal's position.

\section{Backscatter-inertial Super-accuracy State Estimation}
\label{sec:pose}
\begin{figure}[t]
  \centering
  \includegraphics[width=2.5in]{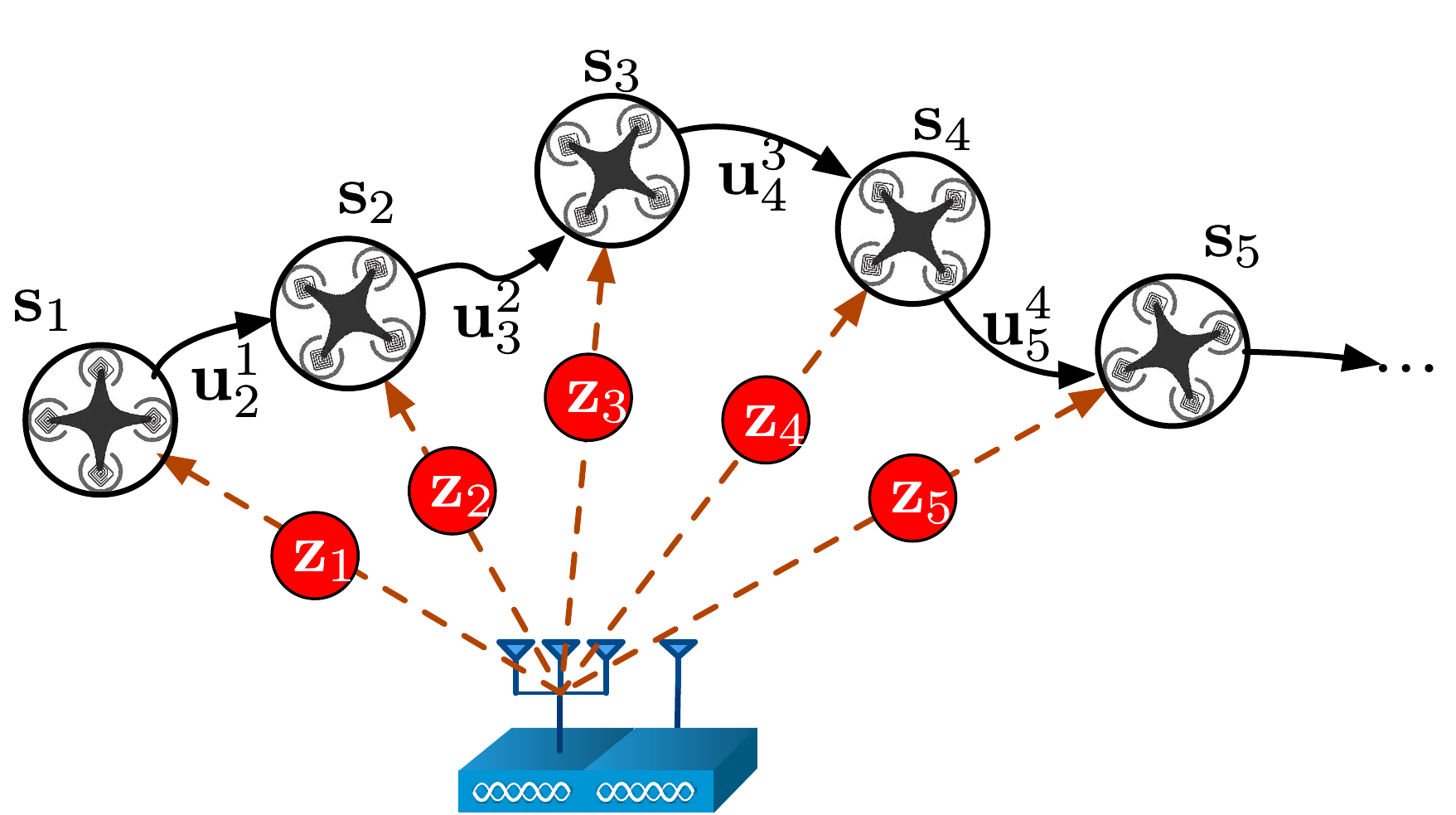}
  \caption{Graph-based optimization.}
  \label{fig:graphslam}
  \vspace{-4mm}
\end{figure}

Basically, solving the state estimation problem consists of estimating the MAV state over its trajectory and the terminal's location. The terminal is essentially a key feature of the environment in which the MAV moves. This falls into the simultaneous localization and mapping (SLAM) problem domain. Solutions to the SLAM problem can be either filtering-based or graph-based approaches. 

We employ a graph-based optimization framework to solve our state estimation problem for better performance~\cite{lin2018autonomous, lu2018simultaneous}. The system is highly nonlinear (refer to Eqn.~\eqref{eqn:backscatter_r} and \eqref{eqn:imu_r}). The key variables that cause the nonlinearity is the rotation $\mathbf{q}$ in difference frames. Thus, an initialization is required (\cref{sec:init}) to provide all necessary values for bootstrapping the subsequent nonlinear optimization solver. Then, we use the Gauss-Newton algorithm to solve the optimization problem. 

\subsection{Problem Formulation}
\label{subsec:formulation}
The graph representation of our state estimation problem is shown in Fig.~\ref{fig:graphslam}. Let $\mathbf{s}_k$ denote the state at time $k$. At each $k$, the MAV observes a set of backscatter sensing measurements $\mathbf{z}_k$ which include range $\hat{d}_k$, angle $\hat{\mathbf{a}}_k \in \mathbb{R}^{3}$ and yaw rotation $\hat{\psi}_k$. $\mathbf{u}_{k+1}^k = \left[\hat{\bm{\alpha}}_{i_{k+1}}^{i_k}; \; \hat{\bm{\beta}}_{i_{k+1}}^{i_k}; \; \hat{\bm{\gamma}}_{i_{k+1}}^{i_k}\right]$ is the preintegrated result over IMU measurements (defined in Eqn.~\eqref{eqn:integration}) that represents the odometry between two consecutive states, \ie, $\mathbf{s}_k$ and $\mathbf{s}_{k+1}$.

To achieve real-time processing, we still employ an incremental state update scheme~\cite{kaess2012isam2} that takes IMU and backscatter-based measurements in a fixed time interval for state estimation. As long as a new state with its backscatter-based measurements is available, our approach works in a sliding window fashion that incorporates the new state and marginalizes the oldest state. The marginalization will follow a similar way to~\cite{zhang2020robot}. It converts the estimated information from the marginalized measurements into a new prior $\{\mathbf{b}_p, \mathbf{H}_p\}$ to constraint later estimates. The full state vector within the window is defined similar to the linear initialization. The difference is two-fold: 1) the extrinsic transformation matrix is included for refinement; 2) the gravity is replaced by rotation $\mathbf{q}_{i_k}^{i_0}$ for combating the IMU drift of rotations.
\begin{equation}
	\begin{aligned}
		\bm{\mathcal{S}} & = \left[ \mathbf{s}_0; \quad \mathbf{s}_1; \quad \cdots; \quad \mathbf{s}_n; \quad \mathbf{s}_b^i; \quad \bm{\rho} \right],	\\
		\mathbf{s}_k & = \left[ \mathbf{p}_{i_k}^{i_0}; \quad \mathbf{v}_{i_k}^{i_k}; \quad \mathbf{q}_{i_k}^{i_0} \right], k \in [1, n], \; \mathbf{s}_b^i = \left[\mathbf{p}_b^i; \quad \mathbf{q}_b^i\right].
	\end{aligned}
\end{equation}

In the state vector, we consider variables in different metrics, \eg, meter for position, m/s for velocity, and radian for orientation. Therefore, we choose the Mahalanobis norm to rescale them with their covariance matrices. The covariance matrices of measurements are required to be updated. The objective is to minimize the sum of the Mahalanobis norm of backscatter sensing and IMU residuals to obtain a maximum a posteriori estimation given the prior converted by the marginalization: 
\begin{equation}
	 \min_{\bm{\mathcal{S}}} \left\{\left(\mathbf{b}_p - \mathbf{H}_p\bm{\mathcal{S}}\right) + \sum_{j\in\mathcal{L}}\left\|\mathbf{e}_{\mathcal{L}}\left( \hat{\mathbf{z}}_j, \bm{\mathcal{S}} \right) \right\|_{\mathbf{P}_j}^2 + \sum_{k\in\mathcal{I}}\left\|\mathbf{e}_\mathcal{I}\left(\hat{\mathbf{u}}_{k+1}^k, \bm{\mathcal{S}} \right) \right\|_{\mathbf{P}_{k+1}^k}^2\right\},
	 \label{eqn:nonlinear}
\end{equation}
where $\mathbf{e}_{\mathcal{L}}\left( \hat{\mathbf{z}}_j, \bm{\mathcal{S}} \right)$ (briefly denoted as $\mathbf{e}_{\mathcal{L}}^j$) and $\mathbf{e}_\mathcal{I}\left(\hat{\mathbf{u}}_{k+1}^k, \bm{\mathcal{S}} \right)$ (briefly denoted as $\mathbf{e}_\mathcal{I}^k$) are measurement residuals for LoRa backscatter and IMU, respectively. 

\subsection{Backscatter-inertial State Estimation}
\label{subsec:estimation}
We now solve the nonlinear system~\eqref{eqn:nonlinear} for state estimation via the Gauss-Newton algorithm. This involves linearizing the nonlinear system by the first order Taylor expansion of the residuals in~\eqref{eqn:nonlinear} around the initial values provided by \cref{sec:init}, \ie, computing Jacobians. We can use Ceres Solver~\cite{ceres-solver}, which is an open-source C++ library that solves complicated optimization problems, to automatically compute these complicated Jacobians and solve~\eqref{eqn:nonlinear} to obtain the state estimates. To use this tool, we have to define the measurement residuals for creating template functors. 

{\bf Backscatter sensing residual}. 
Given range $\hat{d}_{b_j}$, angle $\hat{\mathbf{a}}_{b_j}$, and rotation $\hat{\mathbf{q}}_{b_j}^{b_0}$, the residual is defined as, 
\begin{equation}
	\mathbf{e}_{\mathcal{L}}^j = 
	\begin{bmatrix}
		\delta d_{b_j}					\\
		\delta\mathbf{a}_{b_j}	\\
		\delta\bm{\theta}_{b_j}
	\end{bmatrix} = 
	\begin{bmatrix}
		\left\|\hat{d}_{b_j}^2 -  \left( \mathbf{p}_{b_j}^{b_0} - \bm{\rho} \right)^\top \left( \mathbf{p}_{b_j}^{b_0} - \bm{\rho} \right)\right\|	\\
		\hat{\mathbf{a}}_{b_j} \times \left( \mathbf{p}_{b_j}^{b_0} - \bm{\rho} \right)	\\
		2\left[\inv{(\hat{\mathbf{q}}_{b_j}^{b_0})}\otimes \mathbf{q}_{b_j}^{b_0} \right]_{xyz}
	\end{bmatrix},
	\label{eqn:backscatter_r}
\end{equation}
where $[\cdot]_{xyz}$ extracts the vector part of the quaternion, which is the approximation of the error-state representation. $\delta\bm{\theta}_{b_j}$ is the 3D error-state representation of quaternion. The covariance matrix $\mathbf{P}_{b_j}$ is the measurement noise matrix, which can be estimated by statistically analyzing the pose features.

{\bf IMU residual}. 
Based on the kinematics, the residual of IMU measurements can be defined as,
\begin{equation}
	\begin{aligned}
		\mathbf{e}_\mathcal{I}^k & = 
		\begin{bmatrix}
			\delta\bm{\alpha}_{i_{k+1}}^{i_k}	\\
			\delta\bm{\beta}_{i_{k+1}}^{i_k}	\\
			\delta\bm{\gamma}_{i_{k+1}}^{i_k}
		\end{bmatrix} 
		= 
		\begin{bmatrix}
			R(\mathbf{q}_{i_0}^{i_k})\left( \mathbf{p}_{i_{k+1}}^{i_0} - \mathbf{p}_{i_k}^{i_0} + \frac{1}{2}\mathbf{g}^{i_0}\Delta t_k^2 \right) - \mathbf{v}_{i_k}^{i_k}\Delta t_k - \hat{\bm{\alpha}}_{i_{k+1}}^{i_k}	\\
			R(\mathbf{q}_{i_0}^{i_k})\left( R(\mathbf{q}_{i_{k+1}}^{i_0})\mathbf{v}_{i_{k+1}}^{i_{k+1}} + \mathbf{g}^{i_0}\Delta t_k \right) - \mathbf{v}_{i_k}^{i_k} - \hat{\bm{\beta}}_{i_{k+1}}^{i_k}	\\
			2\left[ \inv{(\mathbf{q}_{i_k}^{i_0})} \otimes \mathbf{q}_{i_{k+1}}^{i_0} \otimes \inv{(\hat{\bm{\gamma}}_{i_{k+1}}^{i_k})} \right]_{xyz}
		\end{bmatrix},
	\end{aligned}
	\label{eqn:imu_r}
\end{equation}
where $\hat{\bm{\alpha}}_{i_{k+1}}^{i_k}$, $\hat{\bm{\beta}}_{i_{k+1}}^{i_k}$, and $\hat{\bm{\gamma}}_{i_{k+1}}^{i_k}$ are the preintegrated result defined in Eqn.~\eqref{eqn:integration}. The covariance $\mathbf{P}_{i_{k+1}}^{i_k}$ can be computed in the same way as in \cref{subsec:init}. At this stage, the residuals of the nonlinear system~\eqref{eqn:nonlinear} have been explicitly defined. We next define the error-state representation~\cite{leutenegger2015keyframe} of our system to clarify the linearization process. 

The residuals of the Euclidean part in the state vector such as vehicle's position, velocity, and terminal's position can be written as
\begin{equation}
	\mathbf{p} = \hat{\mathbf{p}} + \delta\mathbf{p}, \quad \mathbf{v} = \hat{\mathbf{v}} + \delta\mathbf{v}, \quad \bm{\rho} = \hat{\bm{\rho}} + \delta\bm{\rho}.
\end{equation}

Since the rotation is non-Euclidean, its residual is modeled as the perturbation in the tangent space of the rotation manifold,
\begin{equation}
	\mathbf{q} = \hat{\mathbf{q}} \otimes \delta\mathbf{q}, \quad \delta\mathbf{q} \approx 
	\begin{bmatrix}
		\frac{1}{2}\delta\bm{\theta}	\\
		1
	\end{bmatrix},
\end{equation}
where $\delta\bm{\theta}$ is the minimal presentation of rotation residual. Thus the full error-state vector can be written as
\begin{equation}
	\begin{aligned}
		\delta\bm{\mathcal{S}} & = \left[ \delta\mathbf{s}_0; \quad \delta\mathbf{s}_1; \quad \cdots; \quad \delta\mathbf{s}_n; \quad \delta\mathbf{s}_b^i; \quad \delta\bm{\rho} \right],	\\
		\delta\mathbf{s}_k & = \left[ \delta\mathbf{p}_{i_k}^{i_0}; \quad \delta\mathbf{v}_{i_k}^{i_k}; \quad \delta\bm{\theta}_{i_k}^{i_0} \right], k \in [1, n], \; \delta\mathbf{s}_b^i = \left[\delta\mathbf{p}_b^i; \quad \delta\bm{\theta}_b^i\right].
	\end{aligned}
\end{equation}
In each Gauss-Newton iteration, system~\eqref{eqn:nonlinear} is linearized at the current state estimate $\hat{\bm{\mathcal{S}}}$ with respect to the error-state vector $\delta\bm{\mathcal{S}}$. Taking the derivative of residual $\mathbf{e}_\mathcal{L}^j$ and $\mathbf{e}_\mathcal{I}^k$ with respect to $\delta\bm{\mathcal{S}}$ produces the corresponding Jacobian matrices. Then we can solve the nonlinear system~\eqref{eqn:nonlinear} by Ceres Solver~\cite{ceres-solver}. 

We summarize the super-accuracy state estimation algorithm in Algorithm~\ref{alg:state}. The goal of this algorithm is to continuously estimate the MAV state by solving the nonlinear system~\eqref{eqn:nonlinear} (Line 1). The pseudocode lists the major steps using Ceres Solver. First, we need the initialization point obtained by solving Eqn.~\eqref{eqn:linear} (Line 2). We set the initial state as the current state (Line 3) and create template functors of residuals based on their measurement models (Line 4--6). In the while-true loop, as long as receiving new measurements of the backscatter-based pose sensing, we first carry out the marginalization from~\cite{zhang2020robot} (Line 8--9). Then we update the vehicle's state (Line 11) with the error-state vector obtained by evaluating the residuals in Ceres Solver (Line 10).
\begin{algorithm} 
\caption{Super-accuracy State Estimation}
\label{alg:state}
\begin{algorithmic}[1]
  \STATE Goal: Continuously estimate the vehicle's state by solving Eqn.~\eqref{eqn:nonlinear} using Ceres Solver~\cite{ceres-solver}
  \STATE Given the initial state and extrinsic parameters $\bm{\mathcal{S}_0}$ obtained by solving Eqn.~\eqref{eqn:linear} 
  \STATE Current state $\hat{\bm{\mathcal{S}}} \leftarrow \bm{\mathcal{S}_0}$
  \STATE Create the template functor of the backscatter sensing residual $T_b$ by model~\eqref{eqn:backscatter_r}
  \STATE Create the template functor of the IMU residual $T_i$ by model~\eqref{eqn:imu_r}
  \STATE Create the template functor of the prior $T_p$ by the marginalization method~\cite{zhang2020robot}
  \WHILE {true} 
  		\IF {Receiving new backscatter-based pose features}
  			\STATE Marginalizing the states~\cite{zhang2020robot}
  			\STATE Error-state vector $\delta\bm{\mathcal{S}}$ $\leftarrow$ Evaluating ($T_b$ + $T_i$ + $T_p$) from current state $\hat{\bm{\mathcal{S}}}$
  			\STATE $\hat{\bm{\mathcal{S}}} \leftarrow \hat{\bm{\mathcal{S}}} + \delta\bm{\mathcal{S}}$
  		\ENDIF
  \ENDWHILE
\end{algorithmic}
\end{algorithm}

\section{Implementation and Evaluation}
\label{sec:evaluation}
\subsection{Implementation and Evaluation Methodology}
The terminal is built by two colocated NI USRP-2943 nodes, each with a UBX160 daughterboard. They have four channels to be configured as a data handler with one antenna and a backscatter signal handler with three antennas. The USRPs are driven by a host computer. We configure USRPs to work on $900$ MHz band. Specifically, the data handler sends $500$ KHz bandwidth signals at $902$ MHz center frequency, which is in US902-928MHz ISM band. The backscatter signal handler receives backscattered signals for the channel phase extraction (\cref{subsec:phase}). The three antennas for the backscatter signal handler are mounted to an acrylic pole separated by a distance of $16$ cm. To ease the prototype implementation, we use a Semtech SX1276MB1LAS long-range transceiver driven by the host computer to send phases to another LoRa transceiver on the MAV for pose sensing (\cref{subsec:pose}). The USRP nodes are synchronized using a NI CDA-2990 8 Channel Clock Distribution Accessory, as an external clock. We run the CSS decoding and the channel phase extraction on the terminal. 

\begin{figure}
	\centering
	\begin{minipage}[b]{0.28\textwidth}\centering
		\center
		\includegraphics[width=1\textwidth]{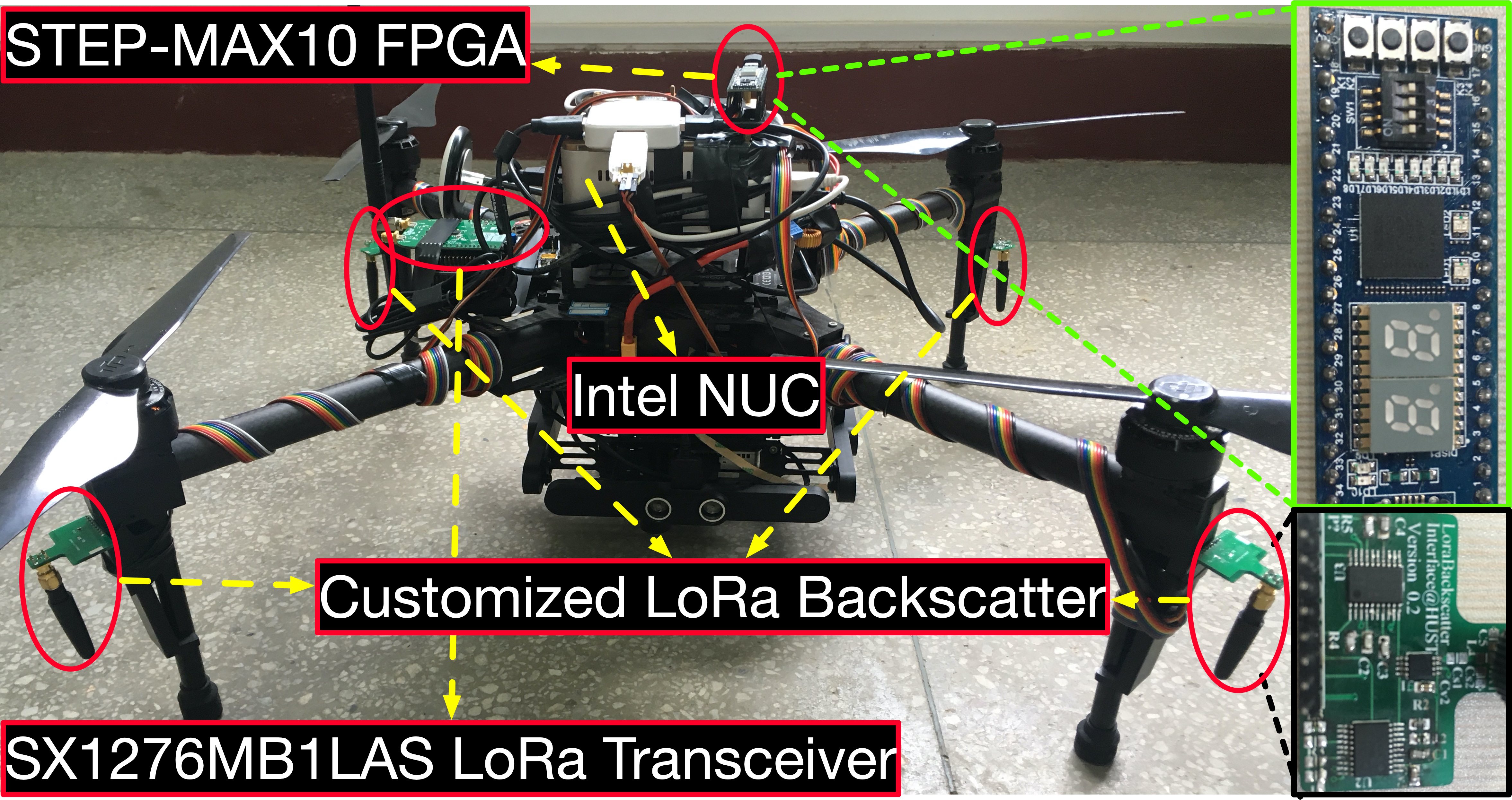}\vspace{-0.3cm}
		\caption{Experiment platform.} \label{fig:sys}
	\end{minipage}
	\hspace{0.1cm}
	\begin{minipage}[b]{0.18\textwidth}\centering
		\center
		\includegraphics[width=1\textwidth]{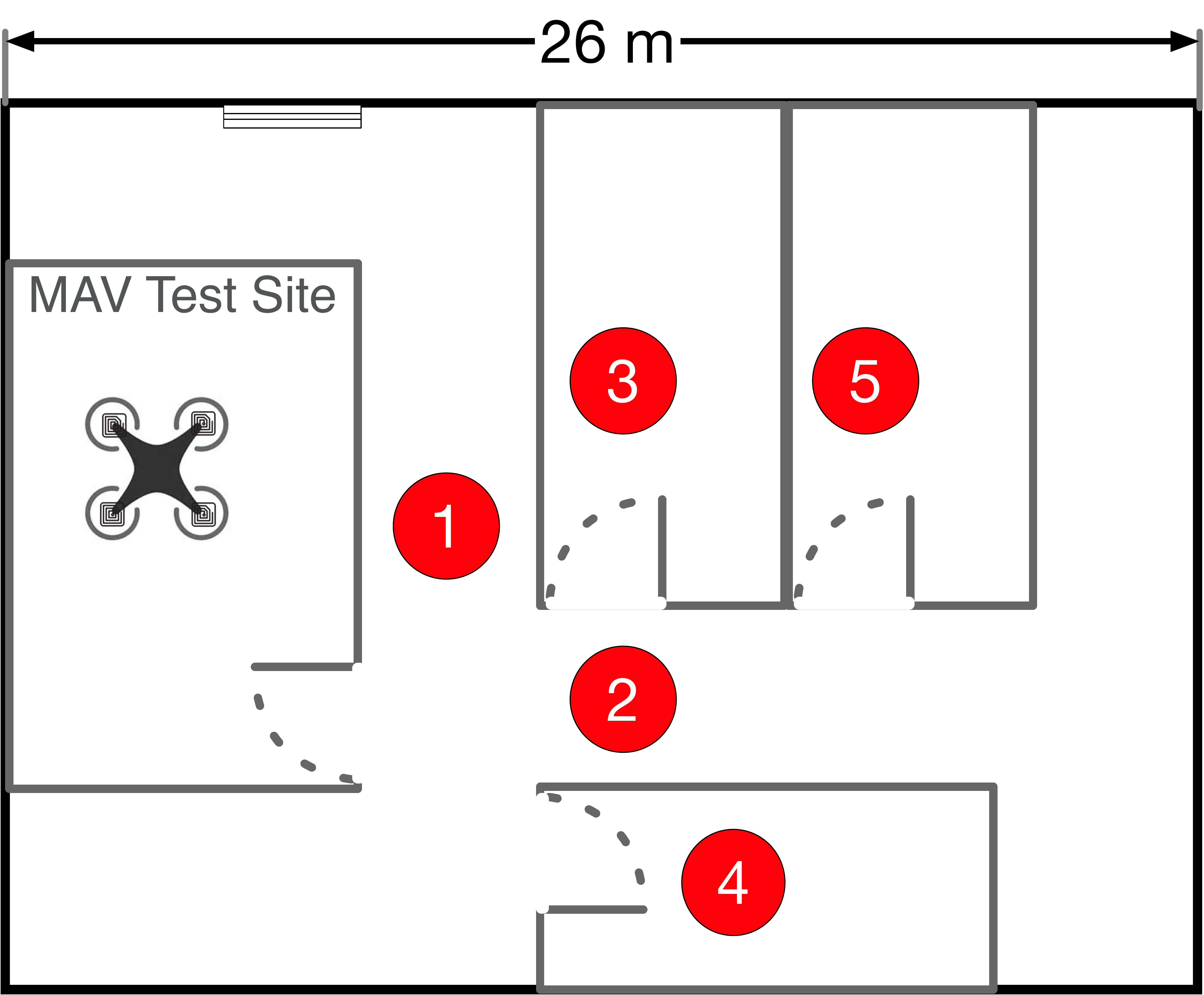}\vspace{-0.3cm}
		\caption{Through-wall setup.} \label{fig:wall_setup}
	\end{minipage}
	\vspace{-4mm}
\end{figure}

The MAV system is built by attaching an Intel NUC, a LORD MicroStrain 3DM-GX4-45 IMU, and an SX1276MB1LAS long-range transceiver on the DJI Matrice 100. In addition, there are four customized LoRa backscatter tags attached on the landing gear of the MAV. The backscatter uses the ADG919 and ADG904 RF switches to enable backscatter communications. The four backscatters are controlled by an Altera STEP-MAX10 FPGA. It configures them to shift $1$ MHz frequency with each other when backscattering the linear chirps with $500$ KHz bandwidth. We run Marvel on the Intel NUC with a $1.3$ GHz Core i5 processor with $4$ cores, an $8$ GB RAM and a $120$ GB SSD, running Ubuntu Linux. The backscatter-based pose sensing module and the backscatter-inertial super-accuracy state estimation algorithm are written in C++. We use Robot Operating System (ROS) to be the interfacing robotics middleware. The experimental platform is shown in Figure~\ref{fig:sys}. All system models and parameters of our experimentations are summarized in Table~\ref{tab:experiment}.
\begin{table}[h]
\footnotesize
  \centering
  \caption{Parameters of our experimentations.}
  \begin{tabular}{ | p{3.6cm} | p{4cm} |}
    \hline
    Component/Configuration & Parameter/Model \\ \hline
    MAV	 platform				& DJI Matrice 100	\\ \hline
    USRP                			& NI USRP-2943 \\ \hline
    Daughterboard 			& UBX160   \\ \hline
    External clock 				& NI CDA-2990   \\ \hline
    LoRa transceiver     	& Semtech SX1276MB1LAS   \\ \hline
    IMU 								& LORD MicroStrain 3DM-GX4-45   \\ \hline
    Onboard computer 		& Intel NUC   \\ \hline
    FPGA							& Altera STEP-MAX10	\\ \hline
    Backscatter switches	& ADG919, ADG904		\\ \hline
    Center frequency of CSS signals	& $902$ MHz	\\ \hline
    Signal bandwidth					& $500$ KHz	\\ \hline
    Frequency shift of tags	& $1$ MHz	\\ \hline
  \end{tabular}
  \label{tab:experiment}
  \vspace{-2mm}
\end{table}

We conduct experiments in both outdoors and indoors for the evaluations in long-range and through-wall settings. The outdoor experiments are conducted in an open field in front of an office building. There is no obstacle between the MAV and the terminal. The indoor experiments are conducted in a MAV test site of $12\times 8$ square meters. The site is located on the basement level of an office building as shown in Fig.~\ref{fig:wall_setup}. Multiple rooms are separated by concrete walls and wooden doors, and have office furniture including tables and computers. To safely conduct indoor experiments, we equipped DJI Guidance~\cite{djiguidance} to detect obstacles. DJI Guidance is a vision-based navigation aid that can perform hovering and obstacle detection in GPS-denied environments. This system will take over the control from Marvel to perform hovering as long as it detects obstacles, \eg, walls and pillars, within $2$ meters of the MAV's surroundings.

\subsection{Micro-benchmark Evaluation}
\label{subsec:microbenchmark}
We evaluate the performance of positioning and rotation estimation, respectively. To evaluate the positioning approach, we build a sliding rail by the stepper motor ROB-09238~\cite{steppermotor} that supports the moving with a controllable speed. We place the MAV on a plate mounted on this rail. To evaluate the rotation estimation, we place the MAV on a plate mounted on the stepper motor and control the rotating speed. In long-range experiments, we place the terminal at one end of the field and move the MAV away from the terminal in increments of $10$ m. In through-wall experiments, we place the MAV in the test site and move the terminal to different rooms (Fig.~\ref{fig:wall_setup}). There are three concrete walls between the terminal and the MAV at location $5$. At each location, we repeat the experiment multiple times and compute the errors. Notice that there are outliers of the backscatter-based pose sensing when we test at position $4$ in Fig.~\ref{fig:wall_setup}. When testing at this position, the doors are open. The MAV occasionally flies near the door and there is no obstacle between the terminal and the MAV at this moment. Therefore, we believe the outliers in this case are due to the change of channel path in the duration of the chirp. Nevertheless, these outliers hardly have negative impacts on the system performance as the majority inliers contribute reliable information to the state estimation, making the optimization subject to the multi-view constraint insensitive to these outliers. 

\begin{figure}
	\centering
	\begin{minipage}[b]{0.24\textwidth}\centering
		\center
		\includegraphics[width=1\textwidth]{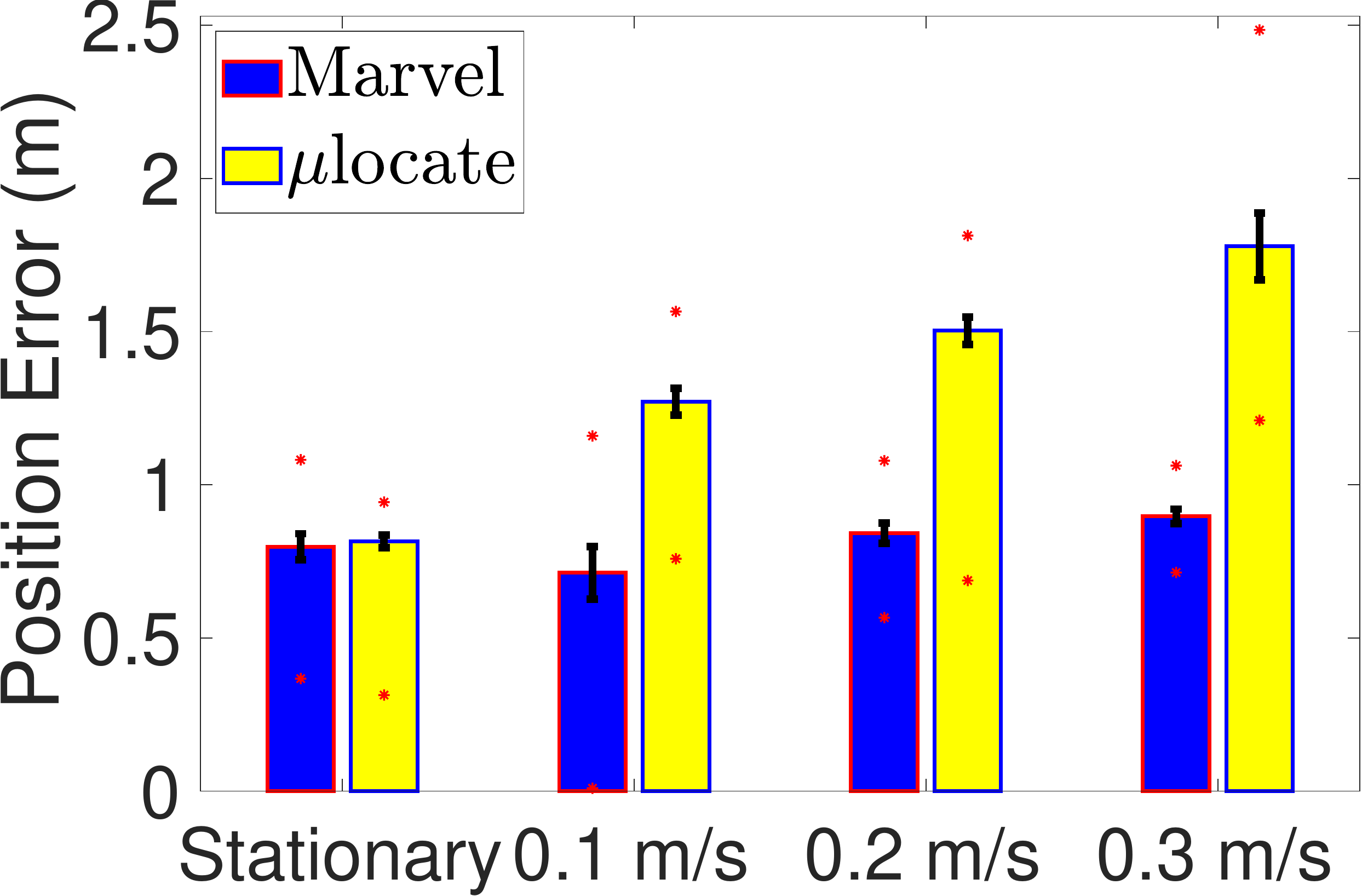}\vspace{-0.3cm}
		\caption{Positioning vs. speed.} \label{fig:comparison}
	\end{minipage}
	\hspace{0.1cm}
	\begin{minipage}[b]{0.22\textwidth}\centering
		\center
		\includegraphics[width=1\textwidth]{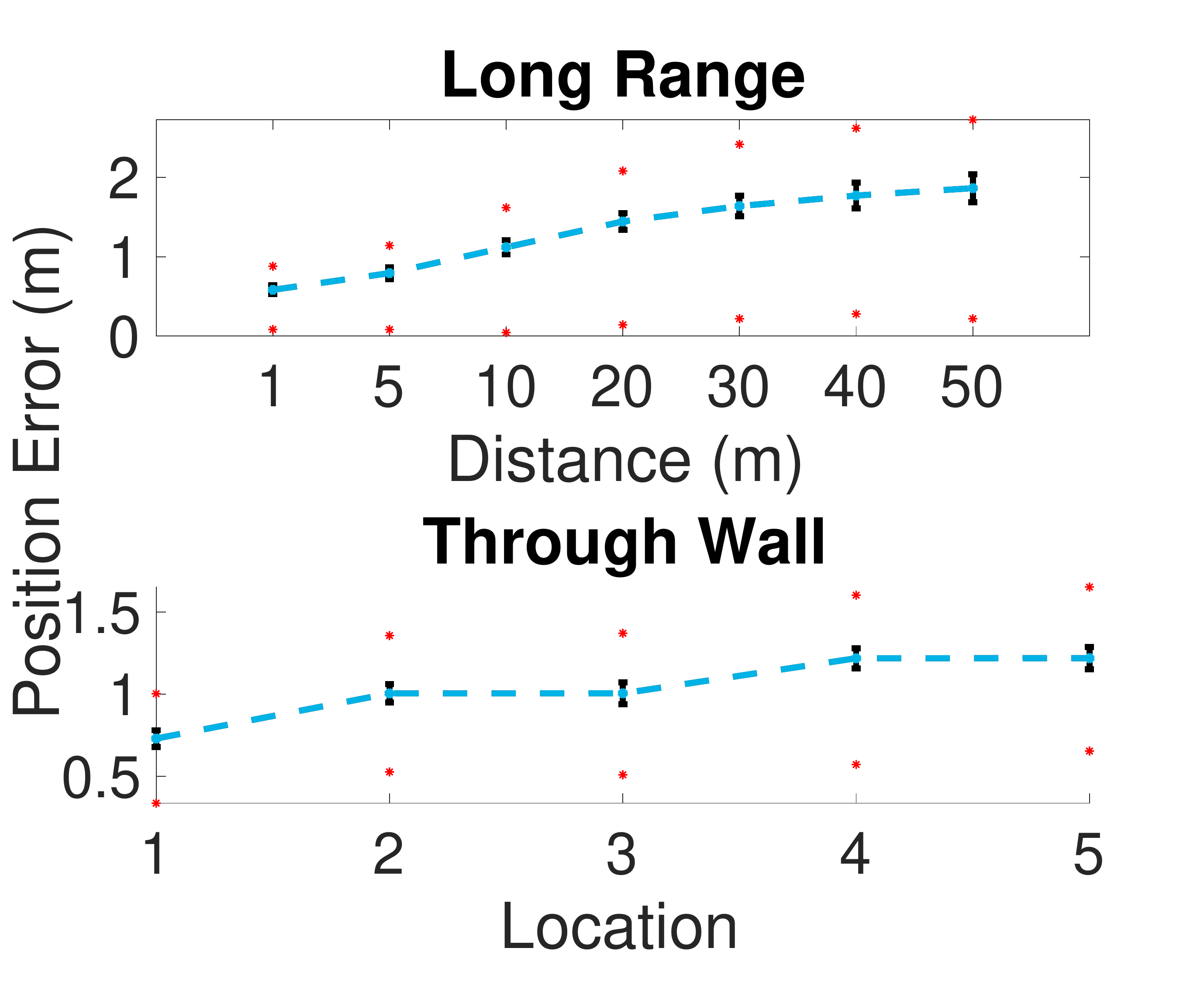}\vspace{-0.3cm}
		\caption{Positioning vs. setting.} \label{fig:position_error}
	\end{minipage}
	\vspace{-4mm}
\end{figure}

{\bf Positioning accuracy}. 
We first validate the positioning capability of Marvel in different speeds. We compare Marvel with the state-of-the-art CSS-based localization system, $\mu$locate~\cite{nandakumar20183d}, which operates correctly in semi-stationary scenarios. As shown in Fig.~\ref{fig:comparison}, the accuracies of the two approaches are similar in stationary case, whose mean error is around $0.8$ m. However, the error of $\mu$locate scales with the speed since its channel phase estimates are distorted by the Doppler frequency shift. The red stars in Fig.~\ref{fig:comparison} denote the best and worst errors over each setting. From the stars, we can see that the worst position error reaches $2.45$ m for $\mu$locate while Marvel's accuracy keeps steady. Meanwhile, we also statistically analyze the results and plot the $95\%$ confidence intervals over the bar. The intervals show that the positioning is quite reliable. In the worst case that runs $\mu$locate at a speed of $0.3$ m/s, the interval is $1.783 \pm 0.109$ m.

The positioning results in different settings are shown in Fig.~\ref{fig:position_error}. The blue dashed lines denote mean errors. The red stars denote the best and worst errors for each setting. We also plot the $95\%$ confidence interval for each setting. To demonstrate that our approach is resilient to the Doppler effect under mobility, we move the MAV along the rail in a speed of $3$ m/s, which is the maximum speed allowed. 

The long-range result shows that the error scales with the MAV-terminal distance. The position error of $0.58$ m at a distance of $1$ m, which increases to $0.79$ m at a distance of $5$ m. This further increases to $1.44$ m at a distance of $20$ m. This is due to the fact that the angle estimate with limited accuracy maps to a growing uncertain area of the MAV's position with the increasing distance. Our customized backscatter works at most $50$ m at which the worst case position accuracy is $2.66$ m. The confidence interval in this case is $1.863 \pm 0.176$ m. Beyond that distance, the power of the received signal is too low to decode even with the CSS coding. 

The through-wall result shows that the accuracies at different locations are similar because the MAV-terminal distance does not vary much. But the accuracy in indoors is worse than at a distance of $1$ m in the open space due to the multipath fading. The worst case accuracy at location $5$ where has three walls blocking the MAV and the terminal is $1.22$ m. The confidence interval in this case is $1.216 \pm 0.067$ m. Our terminal is unable to receive the backscatter signal when it goes through more than three walls. 

In summary, the position accuracy is limited to meter level in both outdoors and indoors due to the limited signal bandwidth at the $900$ MHz band that we use. Nevertheless, with the aid of IMU, Marvel achieves decimeter-level accuracy as shown in \cref{subsec:systemlevel}.

\begin{figure}
	\centering
	\begin{minipage}[b]{0.29\textwidth}\centering
		\center
		\includegraphics[width=1\textwidth]{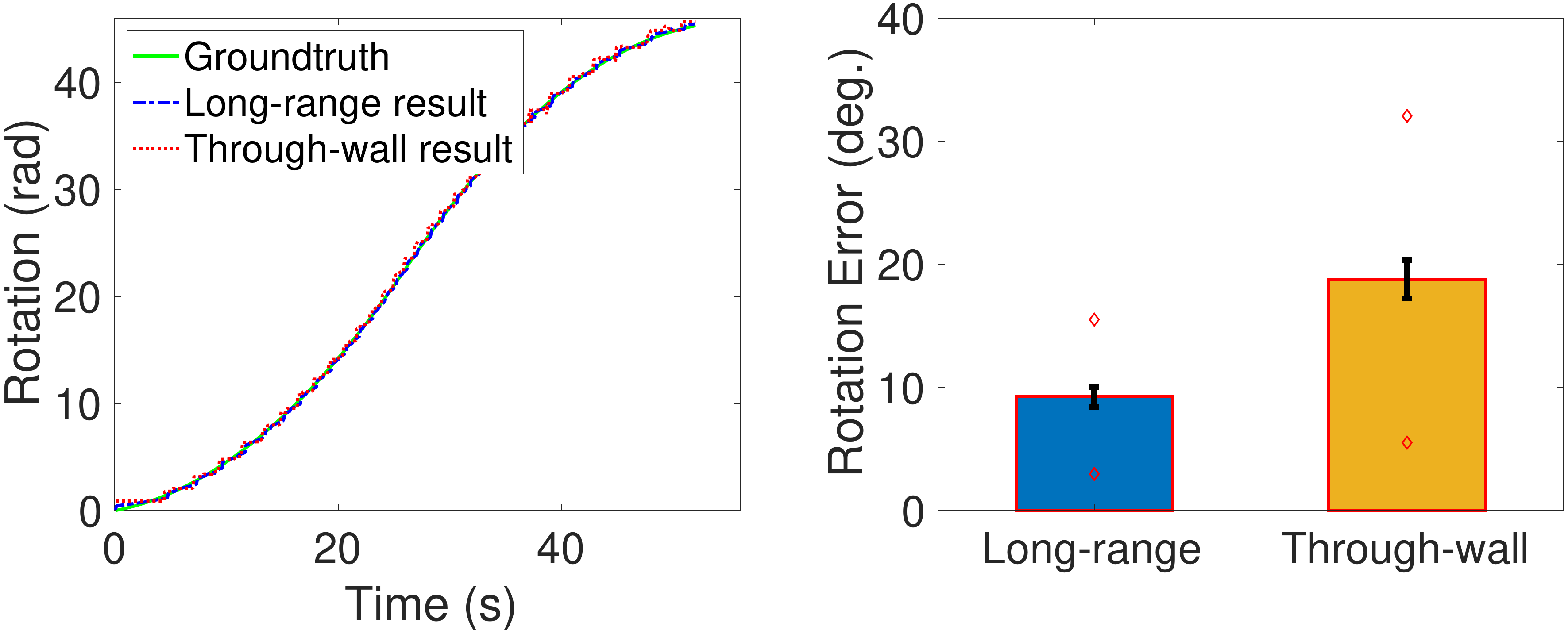}\vspace{-0.3cm}
		\caption{Rotation accuracy.} \label{fig:rotation_accuracy}
	\end{minipage}
	\hspace{0.1cm}
	\begin{minipage}[b]{0.17\textwidth}\centering
		\center
		\includegraphics[width=1\textwidth]{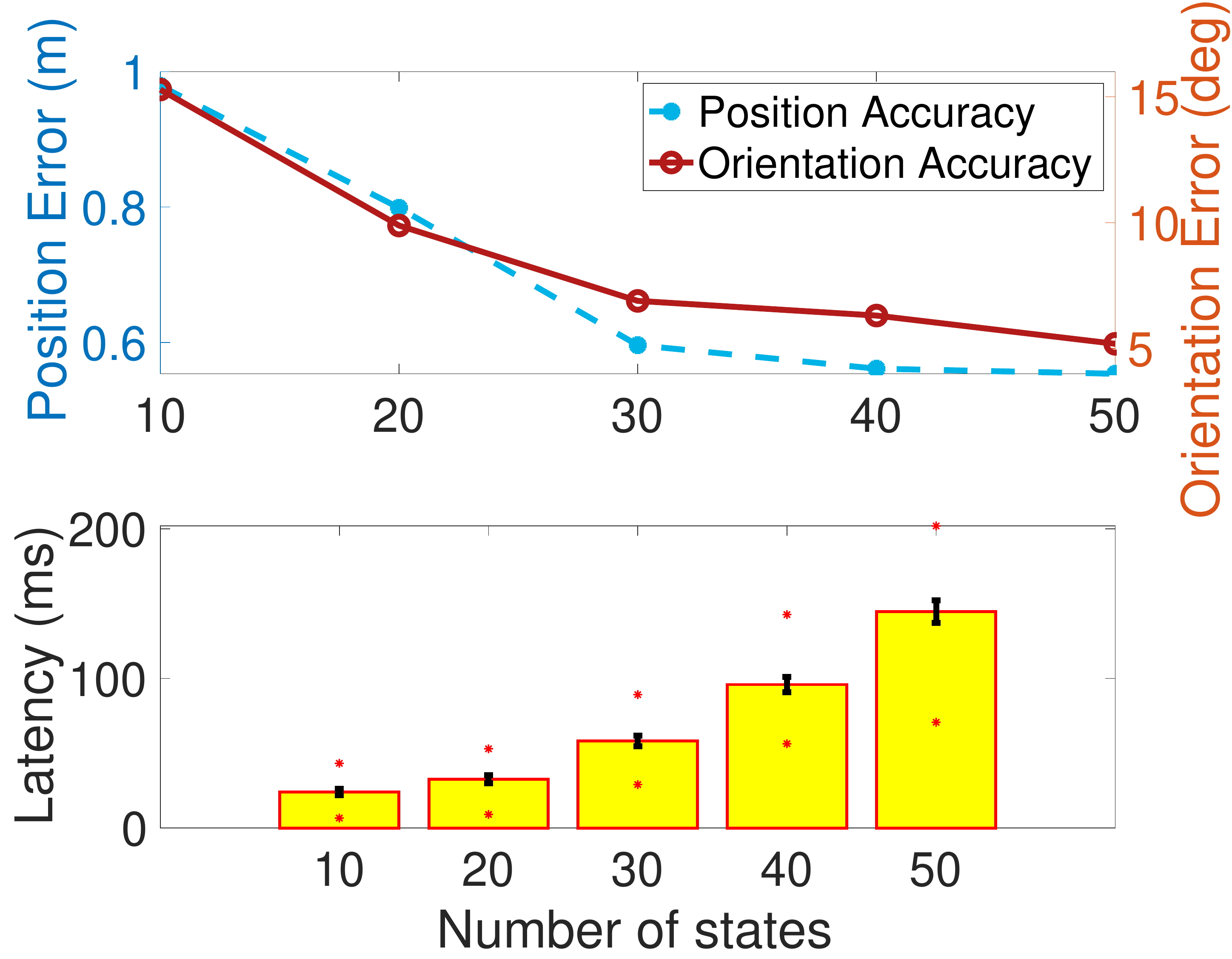}\vspace{-0.3cm}
		\caption{Accuracy vs. latency.} \label{fig:computation_time}
	\end{minipage}
	\vspace{-4mm}
\end{figure}

{\bf Rotation estimation accuracy}. 
We evaluate the rotation estimation by controlling the stepper motor whose angular velocity starts from $0.2$ rad/s and increases by the rate $0.05$ rad/s until $1.5$ rad/s, and then decreases by the same rate to be back at $0.2$ rad/s. The whole process takes $52$ seconds as shown in Fig.~\ref{fig:rotation_accuracy}. We repeat the experiment $60$ times and analyze the data. As expected, the result in the through-wall setting is worse ($95\%$ confidence interval $18.8\degree \pm 1.55\degree$, standard deviation $13.3\degree$) than the other ($95\%$ confidence interval $9.2\degree \pm 0.83\degree$, standard deviation $6.3\degree$) due to the larger error of angle estimation in the presence of multipath. Fig.~\ref{fig:rotation_accuracy} also shows that our rotation estimation algorithm succeeds in closely tracking the MAV's rotation with varying angular velocities in both settings, providing drift-free results.

{\bf Latency}. 
The latency is the key for the real-time property, which is a must for any state estimator for aerial vehicles. The closed form solution (Eqn.~\eqref{eqn:relative_shift}) to the rotation estimation makes the computation efficient. But the other two submodules in Marvel are time-consuming in that the angle estimation requires a parameter search (\cref{subsec:pose}) and the super-accuracy algorithm incorporates a number of states within the optimization framework (\cref{sec:pose}). The computational complexity of the angle estimation is $\mathcal{O}((N_a+ N_t) \times L)$~\cite{kotaru2015spotfi}, where $N_a$ and $N_t$ are the number of steps for each path parameter, $L$ is the number of paths. The computational cost depends on the parameter searching steps and ranges. The super-accuracy algorithm is solved by the Gauss-Newton algorithm, which is an iterative method and has no guaranteed computational complexity in theory. The computational cost depends on the initialization point, the scale of the problem, and the rate of convergence. The whole system is implemented in multiple threads. Thus, the overall system latency depends on the largest time cost between these two submodules. 

On one hand, we test that the average latency of computing an angle is $37.55$ ms. The angle estimation does not hinder the real-time processing. On the other hand, the super-accuracy algorithm has a tradeoff between the accuracy and the latency. The more state involved the more accurate result obtained. But this also increases the latency because a larger state vector and the corresponding measurements are involved in the optimization framework. 
We tune the number of states in the sliding window from $10$ to $50$ for testing. The result in Fig.~\ref{fig:computation_time} shows that when incorporating $50$ states, the positioning accuracy is $0.554$ m, which is only $7\%$ better than the accuracy with $30$ states. However, the $95\%$ confidence interval of latency is $146.088 \pm 7.564$ ms, which is $2.5\times$ slower than the latency with $30$ states. Therefore, we set $30$ states for the rest of our experiments and the average latency is $57.07$ ms for each update. The update rate can reach about $17$ Hz, which is greater than the data rate ($10$ Hz) of the backscatter-based pose sensing, ensuring the real-time processing.

\subsection{System-level State Estimation}
\label{subsec:systemlevel}

\begin{figure}[t!]
    \centering
    \shortstack{
            \includegraphics[width=0.16\textwidth]{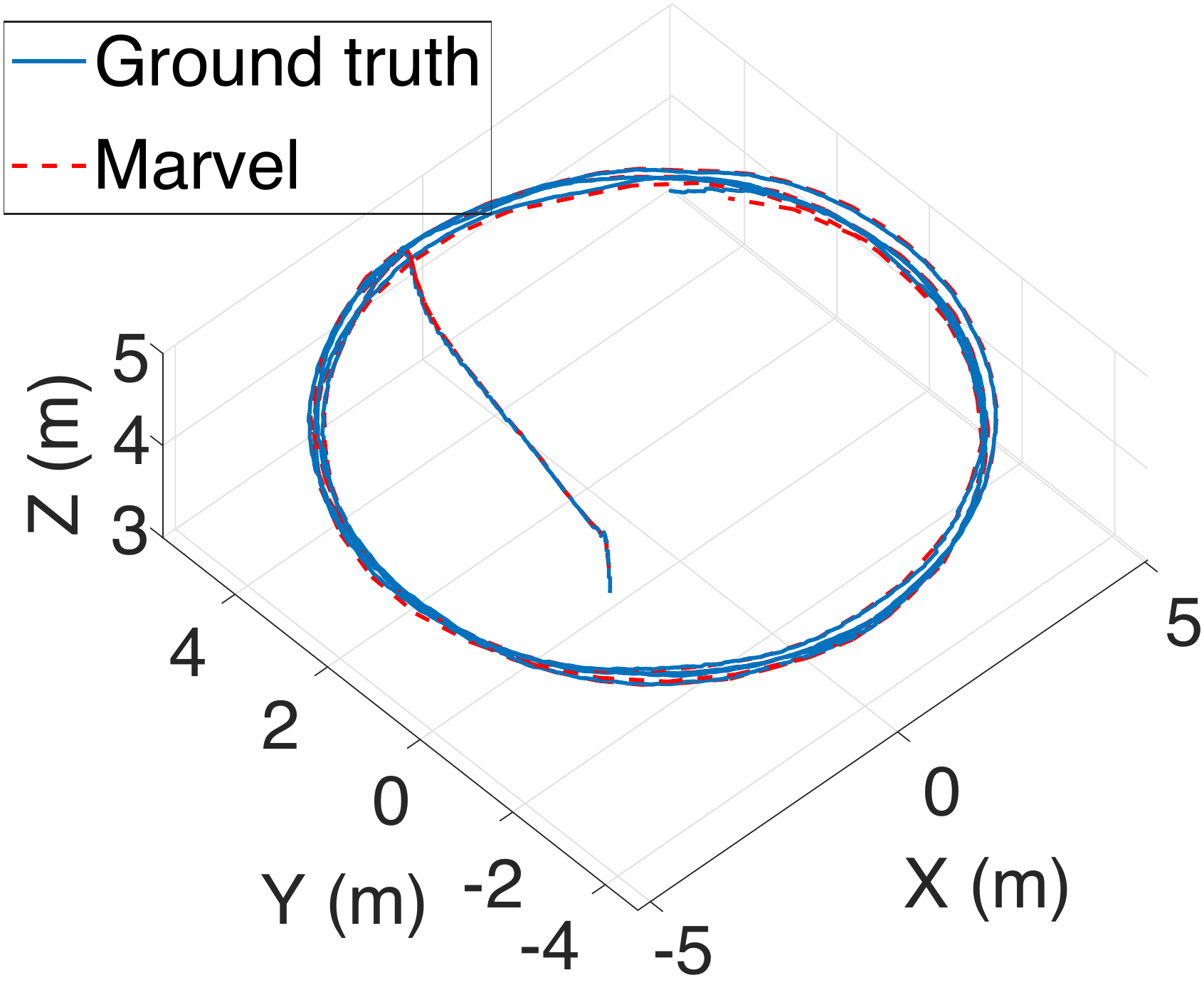}\\
            {\footnotesize (a) Circular trajectory}
    }\quad
    \shortstack{
            \includegraphics[width=0.15\textwidth]{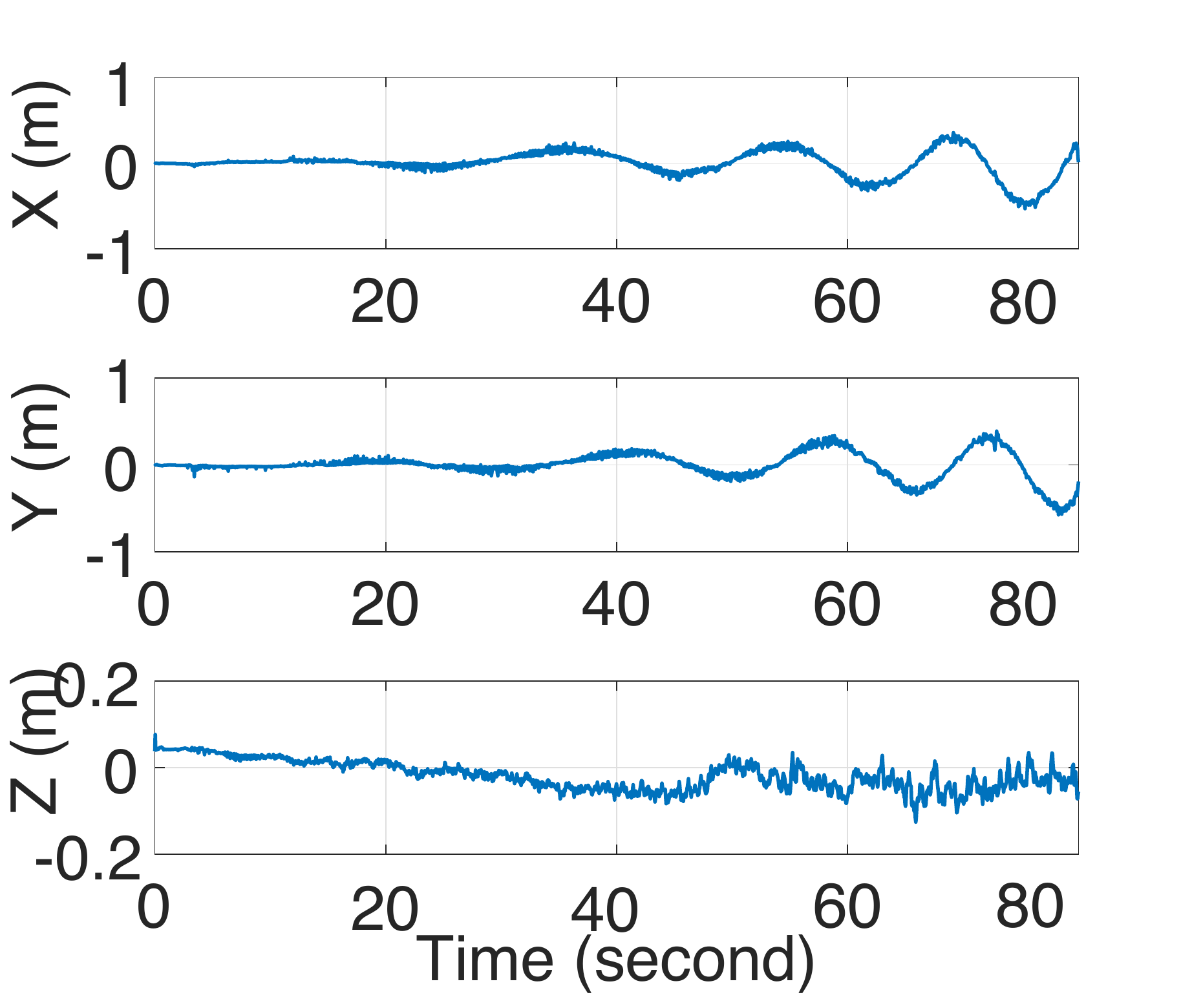}\\
            {\footnotesize (b) Position error}
    }
    \shortstack{
            \includegraphics[width=0.15\textwidth]{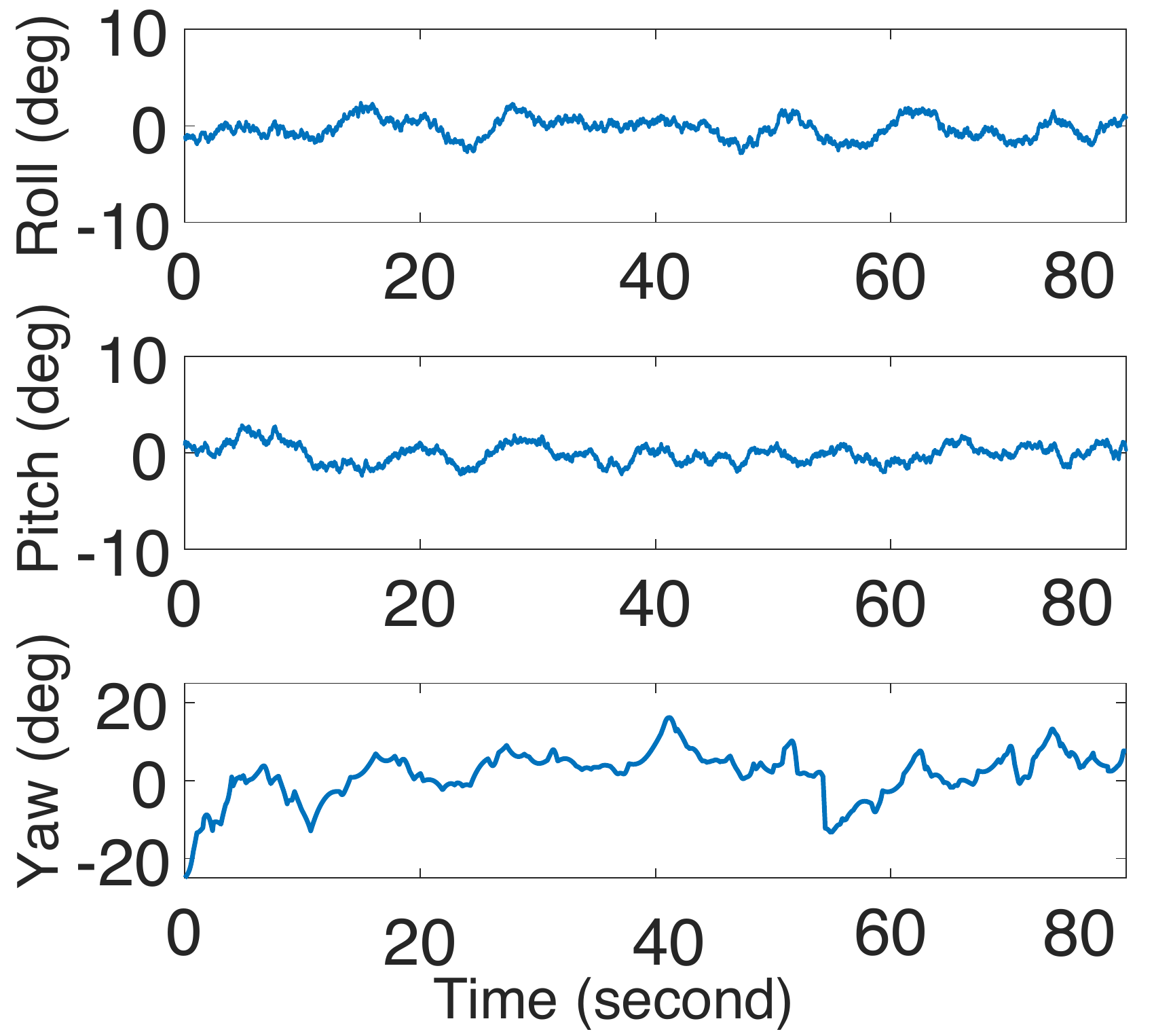}\\
            {\footnotesize (c) Orientation error}
    }
    \caption{Long-range state estimation.}
    \label{fig:losflight} 
    \vspace{-4mm}
\end{figure}

We program the MAV to fly in different trajectories for evaluating the overall performance of Marvel. The ground truth of the flight trajectories is provided by OptiTrack~\cite{optitrack}. The maximum linear velocity reaches $2.53$ m/s in this experiment. 

\subsubsection{Manual initialization and extrinsic calibration}
We first conduct experiments by manually setting the extrinsic parameters, using a vernier caliper and a protractor to measure the relative pose between the IMU and the backscatter sensing. The initial state can be measured by OptiTrack.

In long-range experiments, the MAV flied in a circular trajectory. Since the backscattered signal cannot be decoded when the distance is longer than $50$ m, the terminal is placed $20$ m away from the MAV before taking off to ensure that the MAV cannot go beyond the distance limitation during the flight. As shown in Fig.~\ref{fig:losflight}, the average error of state estimation is $33.66$ cm for positioning and $4.99\degree$ for orientation estimation. This demonstrates that the super-accuracy algorithm significantly increases the accuracy of pose tracking, enabling accurate state estimation. 

\begin{figure}[t!]
    \centering
    \shortstack{
            \includegraphics[width=0.16\textwidth]{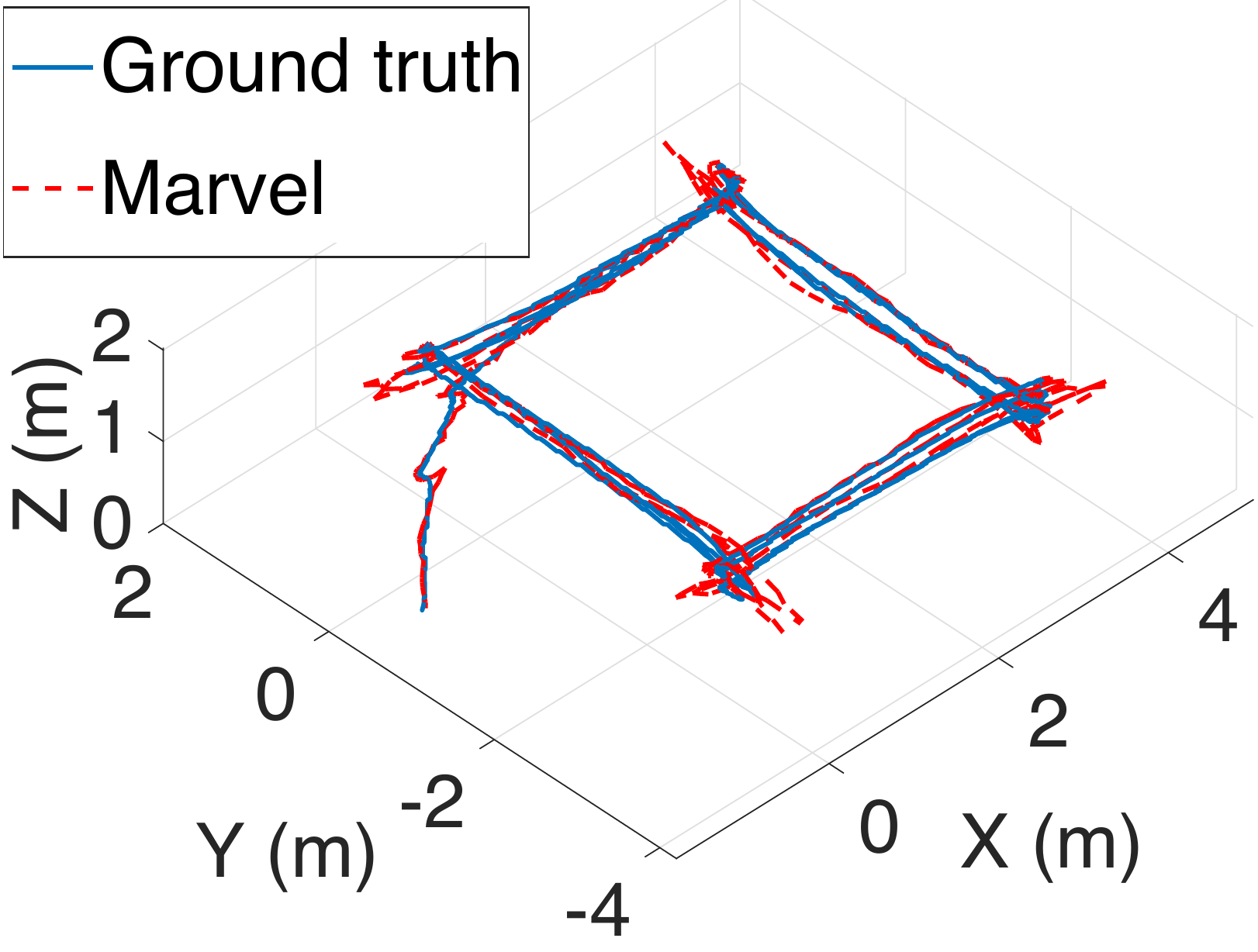}\\
            {\footnotesize (a) Square trajectory}
    }\quad
    \shortstack{
            \includegraphics[width=0.15\textwidth]{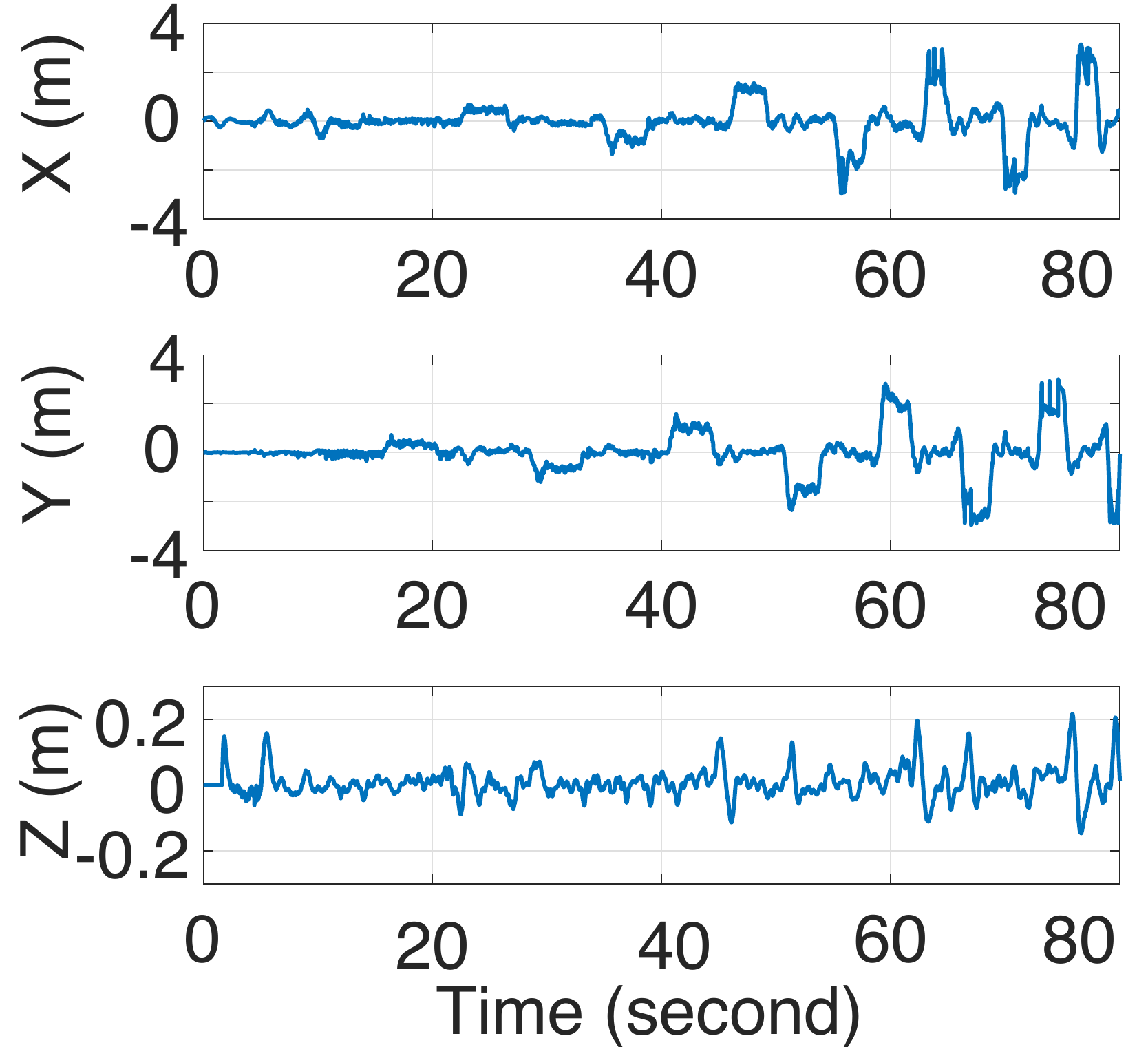}\\
            {\footnotesize (b) Position error}
    }
    \shortstack{
            \includegraphics[width=0.15\textwidth]{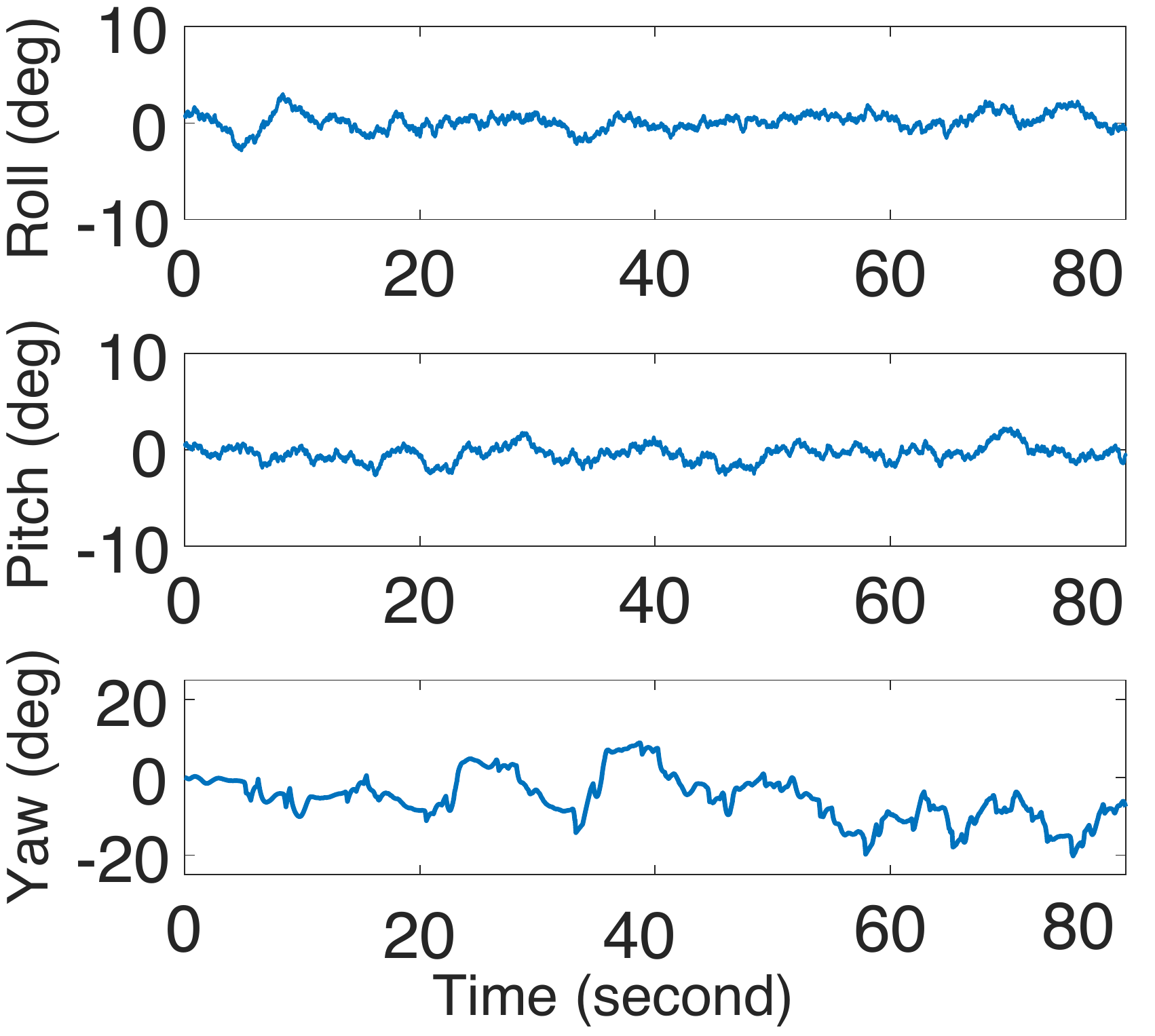}\\
            {\footnotesize (c) Orientation error}
    }
    \caption{Through-wall state estimation.}
    \label{fig:nlosflight}
    \vspace{-4mm}
\end{figure}

In through-wall experiments, for safety reasons, the MAV has to fly in the test site. We placed the terminal at location $5$ and the MAV flied in a square trajectory due to the limited area. As shown in Fig.~\ref{fig:nlosflight}, the average position error over the trajectory is $52.56$ cm and the average orientation error is $6.64\degree$. The accuracy is slightly worse than in the open field due to the multipath fading and the more aggressive motions around the corners of the square trajectory. 

\subsubsection{Online initialization and extrinsic calibration}
We conduct experiments in a similar setting as the previous content except that the online initialization and extrinsic calibration module is working. Initially, we start running the initialization procedure. Then we hold the MAV by hand and move with enough rotations in about $10$ seconds. At this stage, the initialization procedure is completed. The system proceeds to the super-accuracy nonlinear state estimator. 

\begin{figure}[t!]
    \centering
    \shortstack{
            \includegraphics[width=0.16\textwidth]{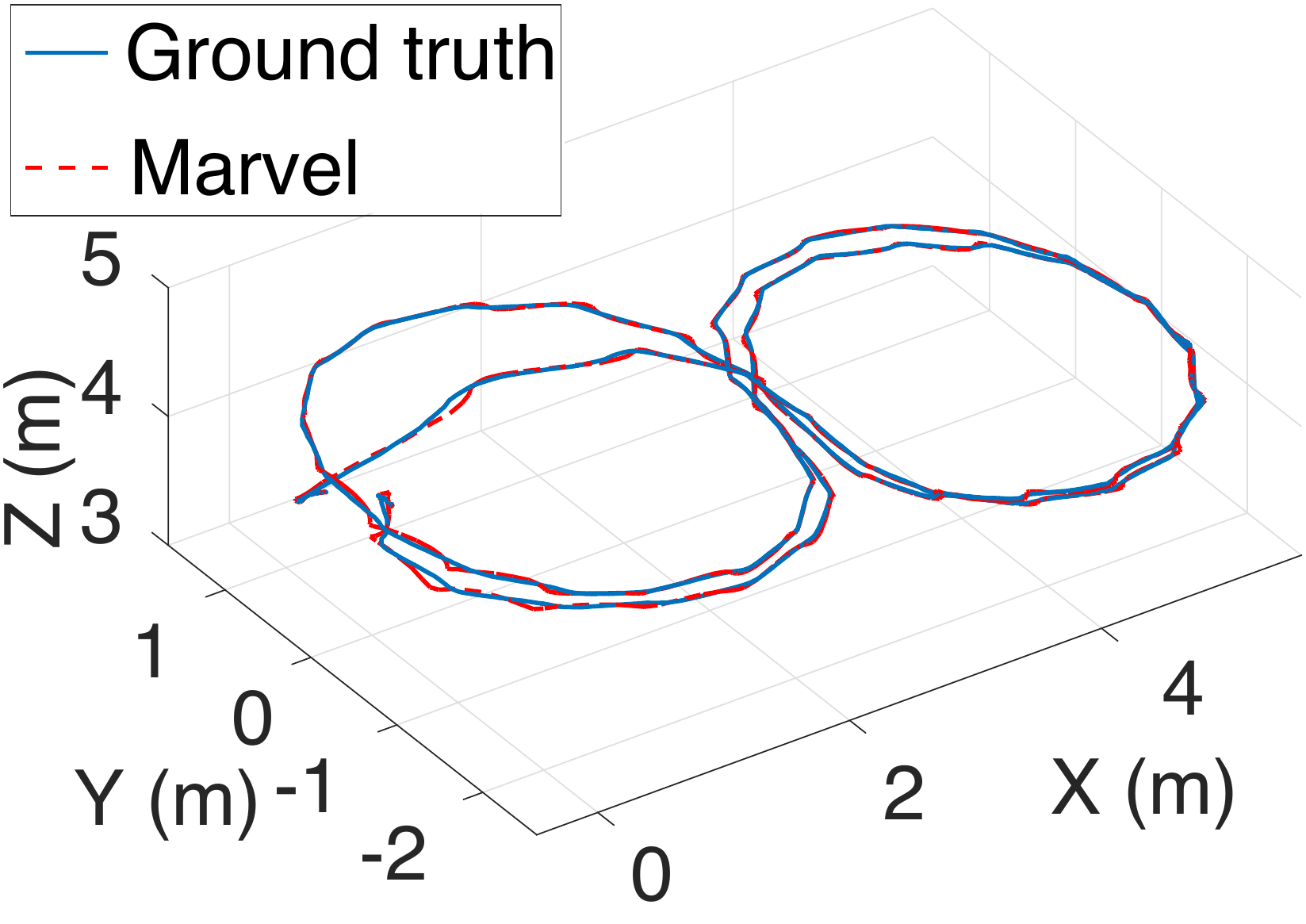}\\
            {\footnotesize (a) Eight-shape trajectory}
    }\quad
    \shortstack{
            \includegraphics[width=0.15\textwidth]{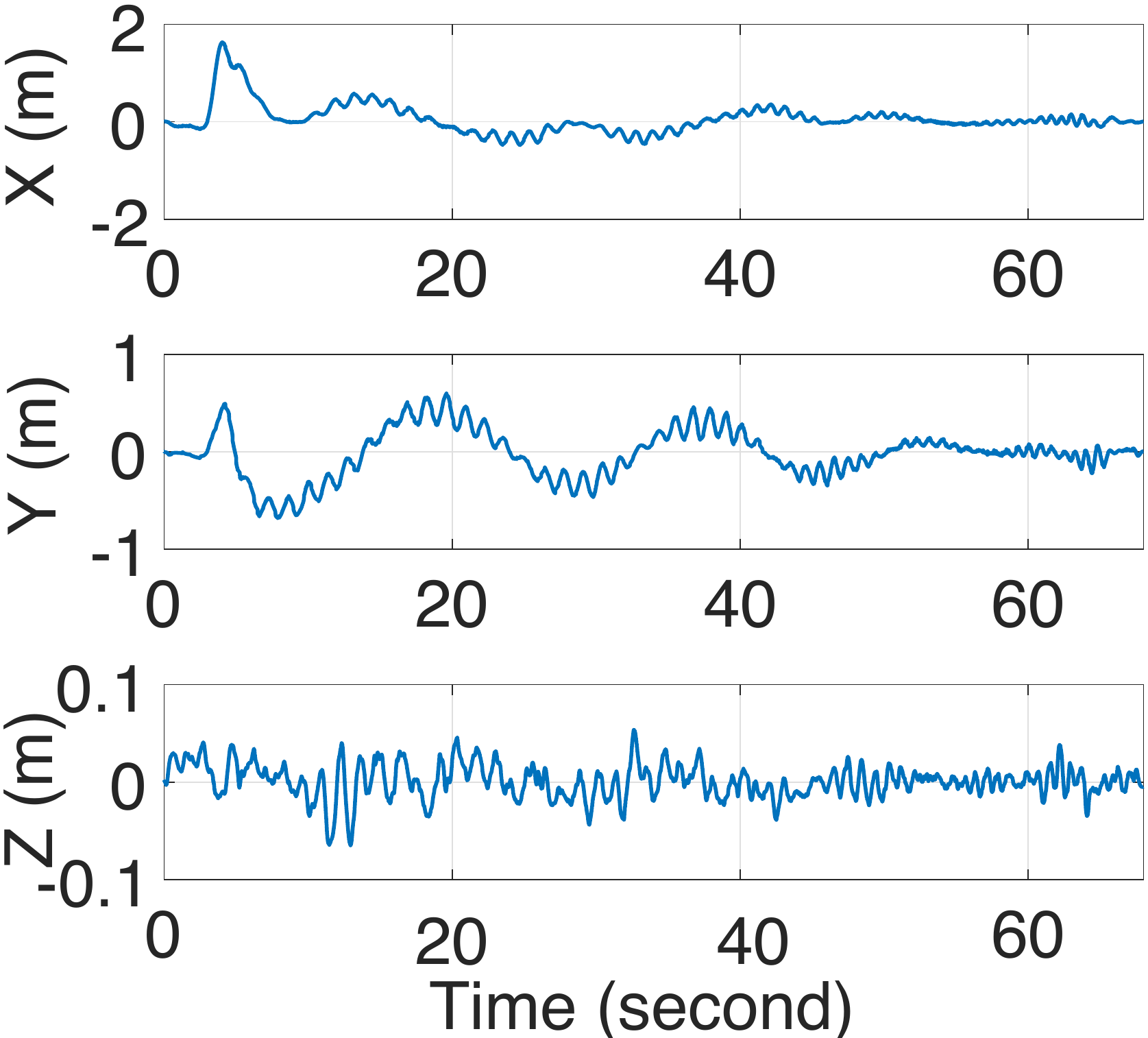}\\
            {\footnotesize (b) Position error}
    }
    \shortstack{
            \includegraphics[width=0.15\textwidth]{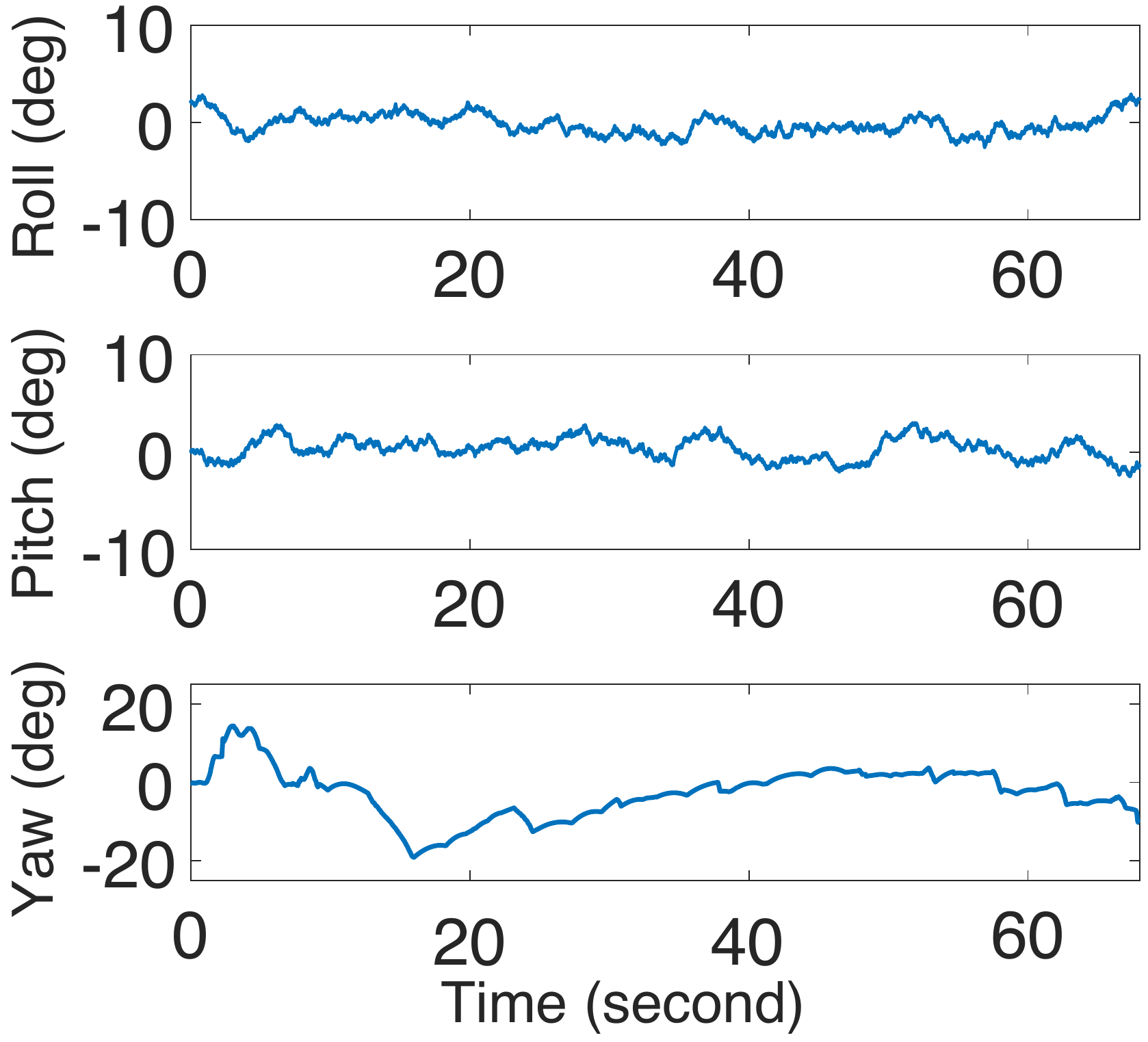}\\
            {\footnotesize (c) Orientation error}
    }
    \caption{Long-range state estimation with online initialization and extrinsic estimation.}
    \label{fig:initialize}
\end{figure}

\begin{figure}[t]
  \centering
  \includegraphics[width=3in]{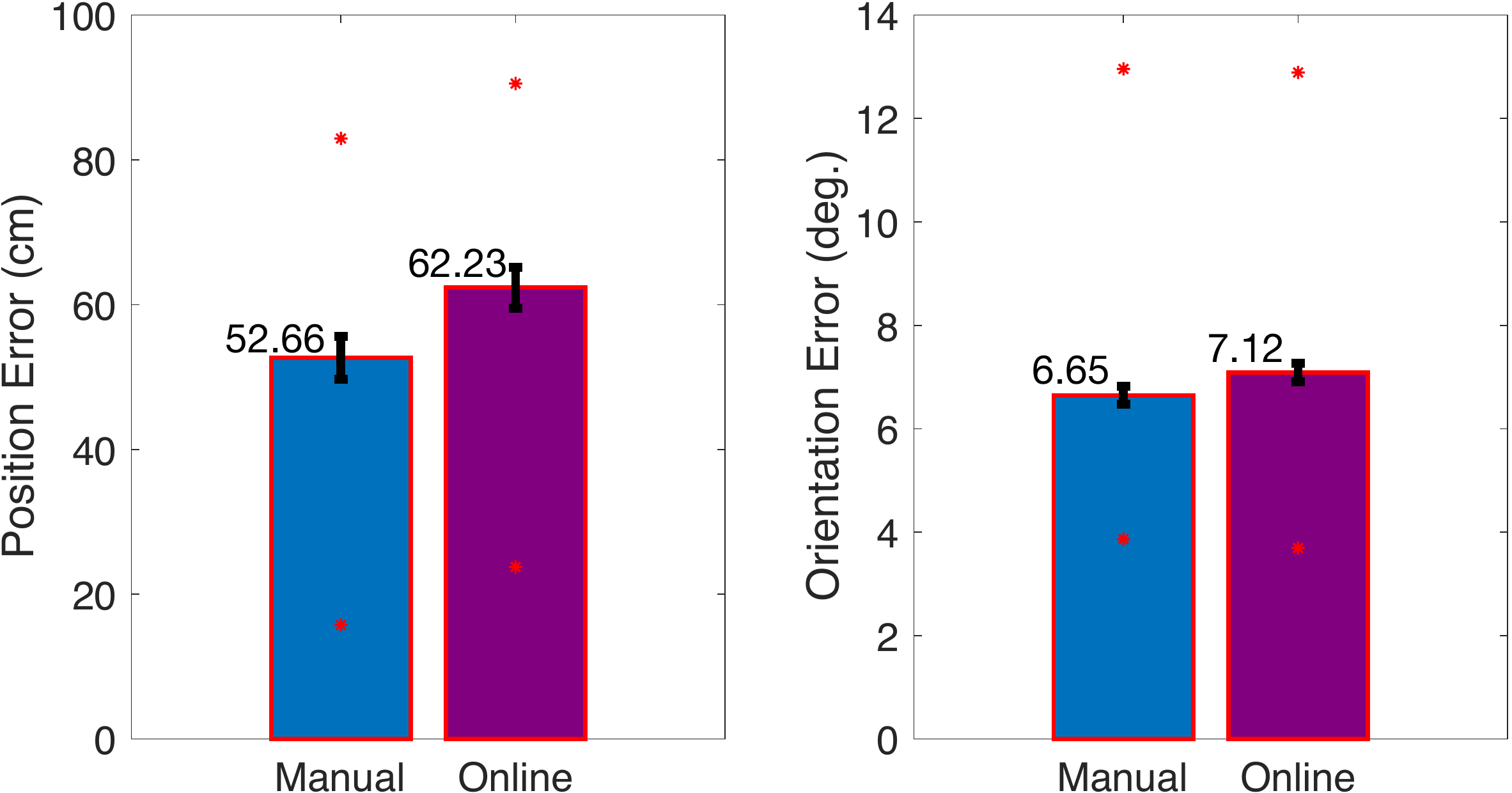}
  \caption{Positioning and orientation estimation accuracies with/without the initialization procedure in the through-wall setting.}
  \label{fig:initialize_throughwall}
  \vspace{-4mm}
\end{figure}

In long-range experiments, we program the MAV flying in an eight-shape trajectory. As shown in Fig.~\ref{fig:initialize}, the average error is $39.18$ cm for positioning and $5.11\degree$ for orientation estimation. The performance is similar to the case with manual settings. In through-wall experiments, the setup and the flying trajectory is the same as the experiments in manual configuration due to safety reasons in the confined area. In the indoor test, we fly the same trajectory as in Fig.~\ref{fig:nlosflight} (the result with the manual solution). If we re-plot the trajectory, they will be highly identical. Only small differences can be carefully found. As a result, we compare the statistical result of positioning and orientation estimation as shown in Fig.~\ref{fig:initialize_throughwall}. The result shows that the online initialization procedure achieves similar performance to the case of manual initialization. The $95\%$ confidence intervals are $62.23 \pm 2.82$ cm for the positioning error and $7.12\degree \pm 0.18\degree$ for the orientation error respectively. 

In principle, the online solution is not as perfect as the manual solution as it optimizes the initial state and extrinsic parameters through noisy sensor measurements. Therefore, Fig.~\ref{fig:initialize_throughwall} exhibits that the online solution is slightly worse than the manual solution. Such a minor difference does not affect the overall performance.

\subsubsection{Comparison with Other RF-based State Estimators}

\begin{figure}[t!]
    \centering
    \shortstack{
            \includegraphics[width=0.23\textwidth]{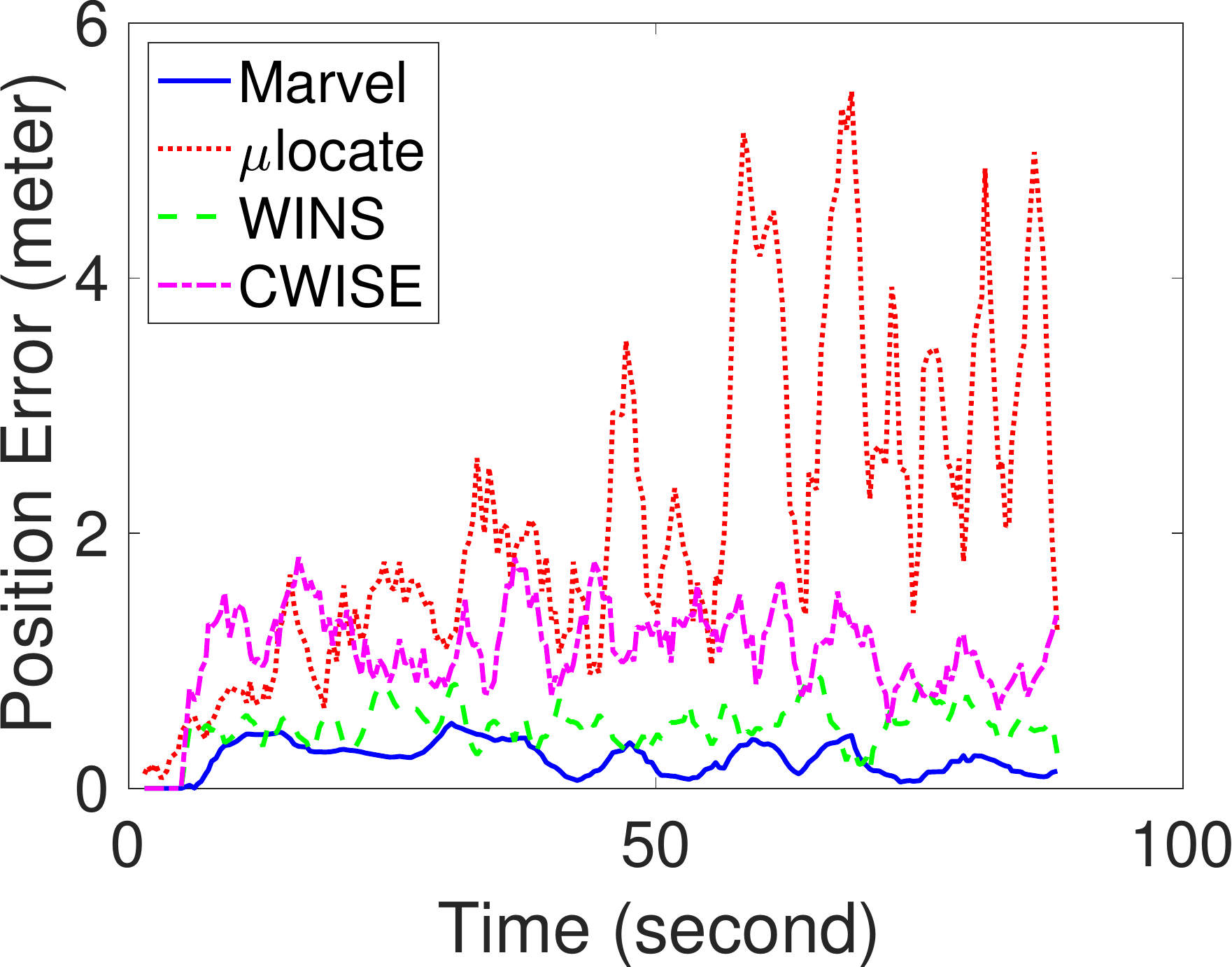}\\
            {\footnotesize (a) Outdoor performance}
    }\quad
    \shortstack{
            \includegraphics[width=0.23\textwidth]{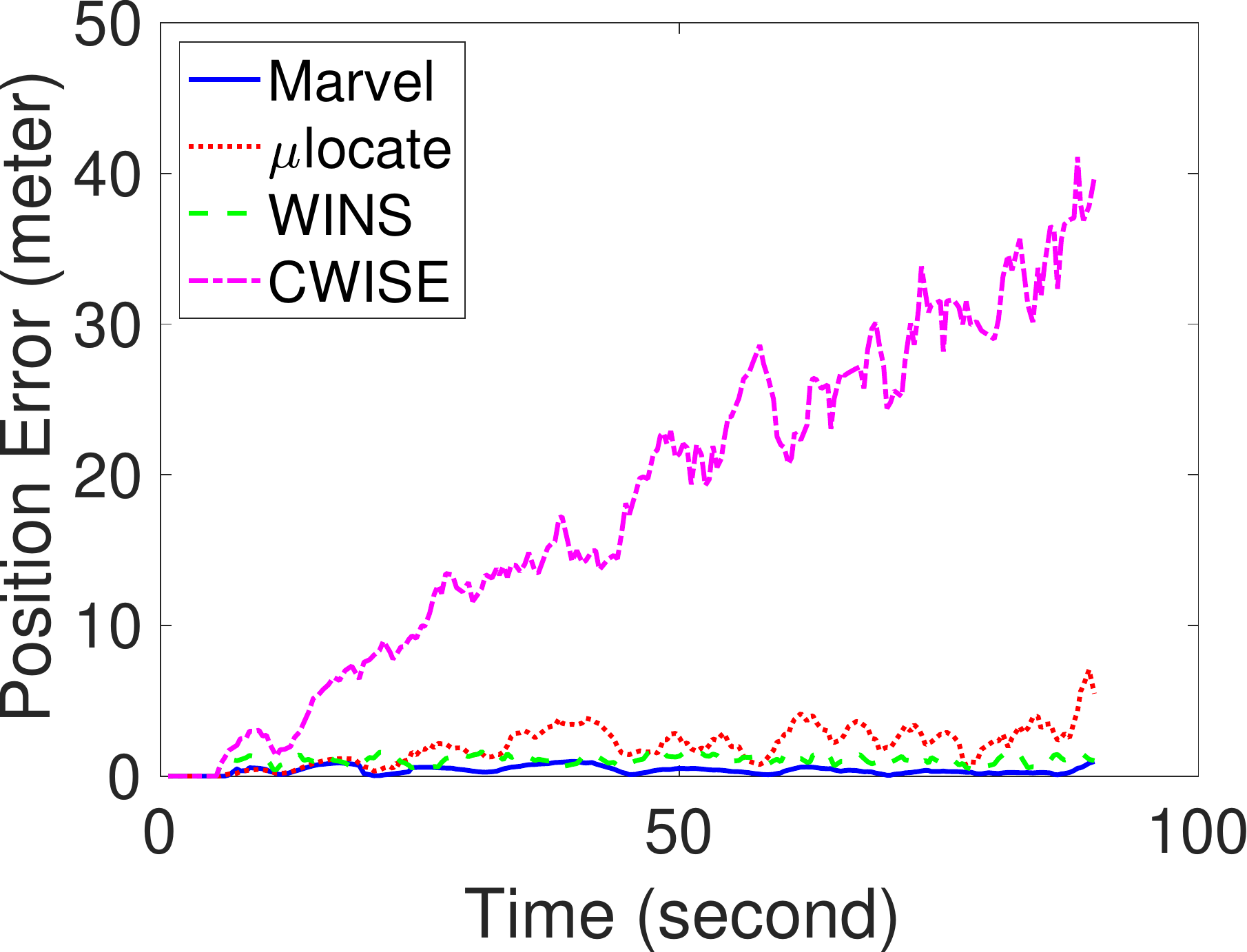}\\
            {\footnotesize (b) Indoor performance}
    }
    \caption{Marvel outperforms the state-of-the-art RF-based state estimators.}
    \label{fig:different_approach}
    \vspace{-4mm}
\end{figure}

In this experiment, we compare Marvel with the state-of-the-art RF-based MAV state estimation systems, CWISE~\cite{li2016csi} and WINS~\cite{zhang2020wifi_full}. They use WiFi signals to estimate MAV states. Therefore, they are incapable of working in a long-range or through-wall setting. To have a fair performance comparison, we place a WiFi access point (AP) at a mild range, in particular, at a distance of $20$ m to the MAV in outdoors. In indoor experiments, we place the AP at location $1$ where one wall blocks the vehicle and the AP. Moreover, we also compare with a modified $\mu$locate~\cite{nandakumar20183d} that fuses the position estimates obtained by $\mu$locate with IMU measurements. Due to the inferior accuracies of other approaches, we still use the state estimates from Marvel to control the vehicle. Then we run other approaches along the same trajectory, in particular, a circular trajectory in outdoors and a squared trajectory in the indoor MAV test site. All experiments are conducted by flying the MAV $90$ seconds. CWISE and WINS are using active radio while Marvel and $\mu$locate uses more challenging RF backscatters to do the state estimation. Since CWISE and $\mu$locate cannot address the orientation estimation, we take the position error to be the evaluation metric.

Fig.~\ref{fig:different_approach} shows that Marvel outperforms all other approaches. In the outdoor scenario, the mean position error of Marvel is $0.235$ m. CWISE's performance is similar to WINS's because the multipath is insignificant. But CWISE (mean error $1.073$ m) is still worse than WINS (mean error $0.525$ m) as there are some reflections of signals from the ground and CWISE is very sensitive to multipath. Marvel benefits from the high sensitivity of LoRa nature against environmental noise, making the phases more faithful to the position. However, $\mu$locate does not eliminate the Doppler effect of fast MAVs, increasing the positioning error as shown in Fig.~\ref{fig:comparison}. Fusing such erroneous estimates with IMU measurements cannot effectively correct the IMU drift. Therefore, its performance is the worst with mean error $2.591$ m and the error tends to be larger along time. 

In the indoor scenario, the mean position error of Marvel is $0.314$ m, being slightly worse due to the presence of multipath. $\mu$locate's performance is similar to the outdoor case because one wall blockage does not affect the decoding capability of LoRa. Its mean error is $2.741$ m. WINS is much worse as expected (mean error $1.056$ m) because the wall's blockage reduces the amplitude of received WiFi signals, making the channel state information (CSI) reported by the WiFi card less accurate and thus increasing the estimation error. Unfortunately, CWISE is completely incapable in indoors due to the multipath. It brings no information to correct the IMU drift. Therefore, we can see that the position error is increasing indefinitely in a fast speed.

\subsection{Discussion}
Here we first briefly discuss the limitation of our system to mapping and path planning. Then, we discuss the energy consumption concern of Marvel.

{\bf Limitations.} The CSS signals we use for state estimation cannot observe obstacles due to its large wavelength since the resolution of environmental observation depends on the wavelength of interactive medium~\cite{kotaru2019light}. Obstacle detection and avoidance relates to mapping and path planning problems, which are also critical to autonomous flight. The system maps the sizes and positions of obstacles and then generates proper trajectories to circumvent obstacles. Optical signals, \eg, visible light captured by cameras, with nanometer-level wavelength are more effective to detect obstacles. However, cameras fail to work in the harsh environment with smoke and fog in our context. Single-chip millimeter wave (mmWave) radar can penetrate airborne obscurants~\cite{lu2020see}, being robust to detect obstacles behind smoke and fog.

{\bf Energy consumption.} Marvel uses low-power backscatters to enable robust state estimation in long-rang or through-wall settings. However, it cannot reduce the energy consumption in our implementation as the additional payload heavily impacts on the energy consumption. Specifically, according to DJI Matrice 100 specifications~\cite{djim100}, the power consumption for hovering is $19$ W per $100$ g. One of our customized backscatter tags weighs $6$ g. Marvel attaches $4$ tags on the landing gear, adding a payload of $24$ g. Thus, the payload additionally consumes $4.56$ W. On the other hand, the power consumption of each backscatter tag is $400$ $\mu$W, four tags on the MAV consumes $1.6$ mW. The power consumption of commercial WiFi is $2.1$ W~\cite{halperin2010demystifying}. In terms of the power consumption of communications, we indeed save energy more than $1000 \times$. However, the total amount of energy saving is about $2.1$ W, which is less than the additional $4.56$ W power consumption for hovering the tags against the gravity. Even we design integrated circuits (ICs) for our backscatter tags, significantly reducing the weight of tags, the power saving for communications is still negligible to the flight. 

According to our test, the DJI M100 can hover $1188$ seconds without adding any hardware component or running any software algorithm. Our prototype adds additional payload with weight about $970$ g, including the expansion bay that can attach additional hardware, an Intel NUC, four backscatter tags, a LoRa transceiver, the DJI guidance system, an FPGA, many TTL cables and a USB expansion. The payload reduces the hovering time from $1188$ to $668$ seconds (reduced by $43.8\%$) without running our algorithms. Running Marvel with the same payload the vehicle can perform hovering $650$ seconds. This concludes that the energy consumption of the software algorithm is negligible compared with the consumed energy from the additional payload. 

Note that our implementation is a prototype that demonstrates the effectiveness of our design. With more advanced hardware, \eg, more compact and powerful DJI Manifold 2, and engineering efforts to miniaturize Marvel's components, \eg, the LoRa transceiver and backscatters, the additional payload can be substantially reduced. In addition, running our software has negligible impacts on the flight time. Therefore, Marvel has no mandatory impact on the MAV flight time.

\section{Related Work}
\label{sec:related}
State estimation for aerial vehicles has been a long studied problem in the robotics community. Before introducing the robot state estimation, we note that there have been numerous research to estimate smartphone's localization/orientation~\cite{yin2017peer, wu2017automatic, zhang2019self, liu2017mercury}. They differ from the robot's state estimation in that robots, \eg, MAVs, require the estimated state to guide their motions, such as hovering, carrying out commands from operators, or navigating to a target position, in unknown environments, while the location and orientation of smartphones are supporting intelligent services for humans other than the devices themselves. Therefore, the location/orientation estimation techniques for smartphones usually do not need to be real-time and robust in a long-term run~\cite{zhou2014use, yin2017peer}, or they are only functional in a known venue with site survey~\cite{wu2017automatic, zhang2019self, liu2017mercury}. On the contrary, the robot's state estimation, especially for MAVs, must be real-time, robust, and accurate in unknown venues. Otherwise, the vehicles can be out of control or even crashed due to that the motor control system takes the erroneous state to produce a thrust in a wrong direction.

{\bf Optical-based state estimation}. LiDAR and camera are the representative sensors to be equipped for state estimation. Dominated vision-based solutions have provided lightweight and accurate state estimates for autonomous MAVs. Noticeable approaches include ORB-SLAM~\cite{mur2017orb}, SVO~\cite{forster2014svo}, and VINS~\cite{qin2017vins}. However, they are limited to well-lighted and texture-rich environments. Although vision-based direct methods~\cite{newcombe2011dtam, pizzoli2014remode} that work with all the raw pixel information in images have better performance in dealing with textureless scenes, they require high computing power (GPUs) to achieve real-time processing, which is unavailable for payload-limited MAVs. In addition, deep learning based approaches~\cite{yin2018geonet, meng2019signet} that learn the mapping between the state and the images are insensitive to light conditions and texture. They either require a labor-intensive site survey to label data for supervised learning, or suffer from inferior performance due to the risk of overfitting.  LiDAR~\cite{droeschel2018efficient, chen2020overlapnet} offers opportunities to address the limitations of vision-based solutions, as it actively illuminates surroundings with laser light. Nevertheless, it is only suitable for standard-size aerial vehicles due to its heavy weight and high cost. Moreover, it also cannot penetrate smoke or fog, making them fail in firefighting operations. 

{\bf RF-based state estimation}. To ease the limitation of optical-based solutions, RF-based state estimators have been proposed in that RF signals are highly resilient to visual limitations. Mueller~\et~\cite{mueller2015fusing} and Liu~\et~\cite{liu2017cooperative} take advantage of UWB-based ranges to enable state estimation. Li~\et~\cite{li2016csi} proposed the first MAV state estimator that leverages WiFi channel state information (CSI), demonstrating the feasibility of state estimation using WiFi. However, it is only functional in outdoors due to the lack of multipath resolution capability. Zhang~\et~\cite{zhang2018wins, zhang2020wifi_full} proposed WINS that addressed the multipath challenge using ubiquitous WiFi. It estimates angle-of-arrivals (AoAs) upon an onboard antenna array and fuses them with IMU measurements to obtain MAV states. 

Recently, communications with low-power signals in long-range or occlusive settings have been studied in~\cite{talla2017lora, peng2018plora}. The signal characteristic and the processing method enable the localizability with such low-power signals. $\mu$locate~\cite{nandakumar20183d} is the first localization system that extracts the channel phases of low-power CSS signals drowned by noise to localize targets by range estimates. It operates correctly in semi-stationary scenarios but not in the presence of agile mobilities of MAVs. The fundamental difference in our context is that the backscattered CSS signals have Doppler frequency shifts. Moreover, $\mu$locate only addresses location and requires the floor plan of the work space to localize targets with a single access point (AP).

\section{Conclusion}
\label{sec:conclusion}
To our knowledge, Marvel represents the first RF backscatter-based MAV state estimation system that works in long range or through wall with online initialization and calibration. It marks a new sensing modality that complements existing visual solutions in supporting MAV navigation. The system is powered by three components: 1) a backscatter-based pose sensing module that estimates pose features via low-power CSS signals; 2) an online initialization and extrinsic calibration module that recover the initial state and extrinsic parameters by simple hand-hold operations; 3) a backscatter-inertial super-accuracy algorithm that leverages IMU for accurate state estimation. We implement Marvel on USRP and the DJI Matrice 100 platform with customized backscatter tags. The experimental results in both outdoors and indoors show that Marvel holds the promise as a long-range/through-wall, lightweight and plug-and-play MAV state estimation system with online initialization and calibration. In future, we plan to seamlessly combine visual sensing and RF sensing to achieve a more robust state estimation system.

\bibliographystyle{IEEEtran}
\bibliography{main}

\begin{thebibliography}{10}
\providecommand{\url}[1]{#1}
\csname url@samestyle\endcsname
\providecommand{\newblock}{\relax}
\providecommand{\bibinfo}[2]{#2}
\providecommand{\BIBentrySTDinterwordspacing}{\spaceskip=0pt\relax}
\providecommand{\BIBentryALTinterwordstretchfactor}{4}
\providecommand{\BIBentryALTinterwordspacing}{\spaceskip=\fontdimen2\font plus
\BIBentryALTinterwordstretchfactor\fontdimen3\font minus
  \fontdimen4\font\relax}
\providecommand{\BIBforeignlanguage}[2]{{%
\expandafter\ifx\csname l@#1\endcsname\relax
\typeout{** WARNING: IEEEtran.bst: No hyphenation pattern has been}%
\typeout{** loaded for the language `#1'. Using the pattern for}%
\typeout{** the default language instead.}%
\else
\language=\csname l@#1\endcsname
\fi
#2}}
\providecommand{\BIBdecl}{\relax}
\BIBdecl

\bibitem{zhang2020rf}
S.~Zhang, W.~Wang, N.~Zhang, and T.~Jiang, ``Rf backscatter-based state
  estimation for micro aerial vehicles,'' in \emph{Proc.~IEEE INFOCOM}, 2020.

\bibitem{lin2018autonomous}
Y.~Lin, F.~Gao, T.~Qin, W.~Gao, T.~Liu, W.~Wu, Z.~Yang, and S.~Shen,
  ``Autonomous aerial navigation using monocular visual-inertial fusion,''
  \emph{J. Field Robot.}, vol.~35, no.~1, pp. 23--51, 2018.

\bibitem{dhekne2019trackio}
A.~Dhekne, A.~Chakraborty, K.~Sundaresan, and S.~Rangarajan, ``Trackio:
  tracking first responders inside-out,'' in \emph{Proc.~USENIX NSDI}, 2019.

\bibitem{guo2019localization}
Q.~Guo, Y.~Zhang, J.~Lloret, B.~Kantarci, and W.~K. Seah, ``A localization
  method avoiding flip ambiguities for micro-uavs with bounded distance
  measurement errors,'' \emph{IEEE Trans. Mobile Comput.}, vol.~18, no.~8, pp.
  1718--1730, 2019.

\bibitem{walmart_drone}
M.~Power, ``Walmart testing warehouse drones to manage inventory,''
  \url{https://www.supplypro.ca/wal-mart-testing-drones-warehouses-manage-inventory/},
  2018, online; accessed 27 July 2019.

\bibitem{firefighter}
P.~L. R.~Fahy and J.~Molis, ``Firefighter fatalities in the united
  states-2017,'' June 2018, {N}ational Fire Protection Association (NFPA)
  Research.

\bibitem{fichtinger2015assessing}
J.~Fichtinger, J.~M. Ries, E.~H. Grosse, and P.~Baker, ``Assessing the
  environmental impact of integrated inventory and warehouse management,''
  \emph{Int. J. Prod. Econ.}, vol. 170, pp. 717--729, 2015.

\bibitem{farrell2008aided}
J.~Farrell, \emph{Aided navigation: {GPS} with high rate sensors}.\hskip 1em
  plus 0.5em minus 0.4em\relax McGraw-Hill, Inc., 2008.

\bibitem{chao2010autopilots}
H.~Chao, Y.~Cao, and Y.~Chen, ``Autopilots for small unmanned aerial vehicles:
  a survey,'' \emph{Int. J. Control. Autom.}, vol.~8, no.~1, pp. 36--44, 2010.

\bibitem{zhu2017event}
A.~Z. Zhu, N.~Atanasov, and K.~Daniilidis, ``Event-based visual inertial
  odometry,'' in \emph{Proc.~IEEE CVPR}, 2017.

\bibitem{dong2019pair}
E.~Dong, J.~Xu, C.~Wu, Y.~Liu, and Z.~Yang, ``Pair-navi: Peer-to-peer indoor
  navigation with mobile visual slam,'' in \emph{Proc.~IEEE INFOCOM}, 2019.

\bibitem{qin2017vins}
T.~Qin, P.~Li, and S.~Shen, ``Vins-mono: A robust and versatile monocular
  visual-inertial state estimator,'' \emph{IEEE Trans. Robot.}, vol.~34, no.~4,
  pp. 1004--1020, 2018.

\bibitem{mur2015orb}
R.~Mur-Artal, J.~M.~M. Montiel, and J.~D. Tardos, ``{ORB-SLAM: a versatile and
  accurate monocular SLAM system},'' \emph{IEEE Trans. Robot.}, vol.~31, no.~5,
  pp. 1147--1163, 2015.

\bibitem{dong2018vinav}
J.~Dong, M.~Noreikis, Y.~Xiao, and A.~Yl{\"a}-J{\"a}{\"a}ski, ``Vinav: A
  vision-based indoor navigation system for smartphones,'' \emph{IEEE Trans.
  Mobile Comput.}, vol.~18, no.~6, pp. 1461--1475, 2018.

\bibitem{liu2017cooperative}
R.~Liu, C.~Yuen, T.-N. Do, D.~Jiao, X.~Liu, and U.-X. Tan, ``Cooperative
  relative positioning of mobile users by fusing imu inertial and uwb ranging
  information,'' in \emph{Proc.~IEEE ICRA}, 2017.

\bibitem{luo20193d}
Z.~Luo, Q.~Zhang, Y.~Ma, M.~Singh, and F.~Adib, ``3d backscatter localization
  for fine-grained robotics,'' in \emph{Proc.~USENIX NSDI}, 2019.

\bibitem{jiang20193d}
C.~Jiang, Y.~He, S.~Yang, J.~Guo, and Y.~Liu, ``3d-omnitrack: 3d tracking with
  cots rfid systems,'' in \emph{Proc.~ACM IPSN}, 2019.

\bibitem{wu2019sigcomm_rim}
C.~Wu, F.~Zhang, Y.~Fan, and K.~J.~R. Liu, ``Rf-based inertial measurement,''
  in \emph{Proc.~ACM SIGCOMM}, 2019.

\bibitem{liu2017mercury}
Z.~Liu, W.~Dai, and M.~Z. Win, ``Mercury: An infrastructure-free system for
  network localization and navigation,'' \emph{IEEE Trans. Mobile Comput.},
  vol.~17, no.~5, pp. 1119--1133, 2017.

\bibitem{zhang2019self}
Z.~Zhang, S.~He, Y.~Shu, and Z.~Shi, ``A self-evolving wifi-based indoor
  navigation system using smartphones,'' \emph{IEEE Trans. Mobile Comput.},
  2019.

\bibitem{mohta2018fast}
K.~Mohta, M.~Watterson, Y.~Mulgaonkar, S.~Liu, C.~Qu, A.~Makineni, K.~Saulnier,
  K.~Sun, A.~Zhu, J.~Delmerico \emph{et~al.}, ``Fast, autonomous flight in
  gps-denied and cluttered environments,'' \emph{Journal of Field Robotics},
  vol.~35, no.~1, pp. 101--120, 2018.

\bibitem{djim100}
``Dji matrice 100,'' \url{https://www.dji.com/matrice100/features}, online;
  accessed 02-September-2020.

\bibitem{sx1276}
SX1276, ``{SX1276/77/78/79 - 137 MHz to 1020 MHz} low power long range
  transceiver,'' \emph{Rev. 5}, 2016.

\bibitem{liando2019known}
J.~C. Liando, A.~Gamage, A.~W. Tengourtius, and M.~Li, ``Known and unknown
  facts of lora: Experiences from a large-scale measurement study,''
  \emph{{ACM} Trans. Sens. Netw.}, vol.~15, no.~2, p.~16, 2019.

\bibitem{note2015loratm}
Modulation, ``{LoRa$^{\text{TM}}$} modulation basics,'' \emph{Semtech, May},
  pp. 1--26, 2015.

\bibitem{nandakumar20183d}
R.~Nandakumar, V.~Iyer, and S.~Gollakota, ``3d localization for sub-centimeter
  sized devices,'' in \emph{Proc.~ACM SenSys}, 2018.

\bibitem{kotaru2015spotfi}
M.~Kotaru, K.~Joshi, D.~Bharadia, and S.~Katti, ``Spotfi: Decimeter level
  localization using wifi,'' in \emph{Proc.~ACM SIGCOMM}, 2015.

\bibitem{trawny2005indirect}
N.~Trawny and S.~I. Roumeliotis, ``Indirect kalman filter for 3d attitude
  estimation,'' \emph{University of Minnesota, Dept. of Comp. Sci. \& Eng.,
  Tech. Rep}, vol.~2, p. 2005, 2005.

\bibitem{shen2015tightly}
S.~Shen, N.~Michael, and V.~Kumar, ``Tightly-coupled monocular visual-inertial
  fusion for autonomous flight of rotorcraft {MAVs},'' in \emph{Proc.~IEEE
  ICRA}, 2015.

\bibitem{lupton2012visual}
T.~Lupton and S.~Sukkarieh, ``Visual-inertial-aided navigation for high-dynamic
  motion in built environments without initial conditions,'' \emph{IEEE Trans.
  Robot.}, vol.~28, no.~1, pp. 61--76, 2012.

\bibitem{kaess2012isam2}
M.~Kaess, H.~Johannsson, R.~Roberts, V.~Ila, J.~J. Leonard, and F.~Dellaert,
  ``isam2: Incremental smoothing and mapping using the bayes tree,'' \emph{Int.
  J. Robot. Res.}, vol.~31, no.~2, pp. 216--235, 2012.

\bibitem{leutenegger2015keyframe}
S.~Leutenegger, S.~Lynen, M.~Bosse, R.~Siegwart, and P.~Furgale,
  ``Keyframe-based visual-inertial odometry using nonlinear optimization,''
  \emph{Int. J. Robot. Res.}, vol.~34, no.~3, pp. 314--334, 2015.

\bibitem{shen2016initialization}
S.~Shen, Y.~Mulgaonkar, N.~Michael, and V.~Kumar, ``Initialization-free
  monocular visual-inertial state estimation with application to autonomous
  mavs,'' in \emph{Experimental robotics}.\hskip 1em plus 0.5em minus
  0.4em\relax Springer, 2016, pp. 211--227.

\bibitem{lu2018simultaneous}
C.~X. Lu, Y.~Li, P.~Zhao, C.~Chen, L.~Xie, H.~Wen, R.~Tan, and N.~Trigoni,
  ``Simultaneous localization and mapping with power network electromagnetic
  field,'' in \emph{Proc.~ACM MobiCom}, 2018.

\bibitem{zhang2020robot}
S.~Zhang, W.~Wang, S.~Tang, S.~Jin, and T.~Jiang, ``Robot-assisted backscatter
  localization for iot applications,'' \emph{IEEE Transactions on Wireless
  Communications}, vol.~19, no.~9, pp. 5807--5818, 2020.

\bibitem{ceres-solver}
S.~Agarwal, K.~Mierle, and Others, ``Ceres solver,''
  \url{http://ceres-solver.org}.

\bibitem{djiguidance}
``{DJI Guidance},'' \url{https://www.dji.com/hk-en/guidance}, online; accessed
  02-January-2021.

\bibitem{steppermotor}
``Stepper motor with cable,'' \url{https://www.sparkfun.com/products/9238}.

\bibitem{optitrack}
``Optitrack -- motion capture systems,'' \url{https://optitrack.com/}.

\bibitem{li2016csi}
B.~Li, S.~Zhang, and S.~Shen, ``{CSI-based WiFi-inertial state estimation},''
  in \emph{Proc.~IEEE MFI}, 2016.

\bibitem{zhang2020wifi_full}
S.~Zhang, W.~Wang, and T.~Jiang, ``Wifi-inertial indoor pose estimation for
  micro aerial vehicles,'' \emph{IEEE Transactions on Industrial Electronics},
  2020.

\bibitem{kotaru2019light}
M.~Kotaru, G.~Satat, R.~Raskar, and S.~Katti, ``Light-field for rf,''
  \emph{arXiv preprint arXiv:1901.03953}, 2019.

\bibitem{lu2020see}
C.~X. Lu, S.~Rosa, P.~Zhao, B.~Wang, C.~Chen, J.~A. Stankovic, N.~Trigoni, and
  A.~Markham, ``See through smoke: robust indoor mapping with low-cost mmwave
  radar,'' in \emph{Proc.~ACM MobiSys}, 2020, pp. 14--27.

\bibitem{halperin2010demystifying}
D.~Halperin, B.~Greenstein, A.~Sheth, and D.~Wetherall, ``Demystifying 802.11n
  power consumption,'' in \emph{Proceedings of International Conference on
  Power Aware Computing and Systems}, 2010, pp. 1--5.

\bibitem{yin2017peer}
Z.~Yin, C.~Wu, Z.~Yang, and Y.~Liu, ``Peer-to-peer indoor navigation using
  smartphones,'' \emph{IEEE JSAC}, vol.~35, no.~5, pp. 1141--1153, 2017.

\bibitem{wu2017automatic}
C.~Wu, Z.~Yang, and C.~Xiao, ``Automatic radio map adaptation for indoor
  localization using smartphones,'' \emph{IEEE Trans. Mobile Comput.}, vol.~17,
  no.~3, pp. 517--528, 2017.

\bibitem{zhou2014use}
P.~Zhou, M.~Li, and G.~Shen, ``Use it free: Instantly knowing your phone
  attitude,'' in \emph{Proc.~ACM MobiCom}, 2014.

\bibitem{mur2017orb}
R.~Mur-Artal and J.~D. Tard{\'o}s, ``Orb-slam2: An open-source slam system for
  monocular, stereo, and rgb-d cameras,'' \emph{IEEE Transactions on Robotics},
  vol.~33, no.~5, pp. 1255--1262, 2017.

\bibitem{forster2014svo}
C.~Forster, M.~Pizzoli, and D.~Scaramuzza, ``Svo: Fast semi-direct monocular
  visual odometry,'' in \emph{Proc.~IEEE ICRA}, 2014, pp. 15--22.

\bibitem{newcombe2011dtam}
R.~A. Newcombe, S.~J. Lovegrove, and A.~J. Davison, ``Dtam: Dense tracking and
  mapping in real-time,'' in \emph{Proc.~IEEE ICCV}, 2011, pp. 2320--2327.

\bibitem{pizzoli2014remode}
M.~Pizzoli, C.~Forster, and D.~Scaramuzza, ``Remode: Probabilistic, monocular
  dense reconstruction in real time,'' in \emph{Proc.~IEEE ICRA}, 2014, pp.
  2609--2616.

\bibitem{yin2018geonet}
Z.~Yin and J.~Shi, ``Geonet: Unsupervised learning of dense depth, optical flow
  and camera pose,'' in \emph{Proc.~IEEE CVPR}, 2018, pp. 1983--1992.

\bibitem{meng2019signet}
Y.~Meng, Y.~Lu, A.~Raj, S.~Sunarjo, R.~Guo, T.~Javidi, G.~Bansal, and
  D.~Bharadia, ``Signet: Semantic instance aided unsupervised 3d geometry
  perception,'' in \emph{Proc~IEEE CVPR}, 2019, pp. 9810--9820.

\bibitem{droeschel2018efficient}
D.~Droeschel and S.~Behnke, ``Efficient continuous-time slam for 3d lidar-based
  online mapping,'' in \emph{Proc.~IEEE ICRA}, 2018, pp. 1--9.

\bibitem{chen2020overlapnet}
X.~Chen, T.~L{\"a}be, A.~Milioto, T.~R{\"o}hling, O.~Vysotska, A.~Haag,
  J.~Behley, C.~Stachniss, and F.~Fraunhofer, ``Overlapnet: Loop closing for
  lidar-based slam,'' in \emph{Proc.~Robotics: Science and Systems (RSS)},
  2020.

\bibitem{mueller2015fusing}
M.~W. Mueller, M.~Hamer, and R.~D'Andrea, ``Fusing ultra-wideband range
  measurements with accelerometers and rate gyroscopes for quadrocopter state
  estimation,'' in \emph{Proc.~IEEE ICRA}, 2015.

\bibitem{zhang2018wins}
S.~Zhang, S.~Tang, W.~Wang, and T.~Jiang, ``Wins: Wifi-inertial indoor state
  estimation for mavs,'' in \emph{Proc.~ACM SenSys}, 2018.

\bibitem{talla2017lora}
V.~Talla, M.~Hessar, B.~Kellogg, A.~Najafi, J.~R. Smith, and S.~Gollakota,
  ``Lora backscatter: Enabling the vision of ubiquitous connectivity,''
  \emph{Proc.~ACM UbiComp}, vol.~1, no.~3, p. 105, 2017.

\bibitem{peng2018plora}
Y.~Peng, L.~Shangguan, Y.~Hu, Y.~Qian, X.~Lin, X.~Chen, D.~Fang, and
  K.~Jamieson, ``{PLoRa}: a passive long-range data network from ambient lora
  transmissions,'' in \emph{Proc.~ACM SIGCOMM}, 2018.

\end{thebibliography}

\end{document}